\documentclass[journal]{IEEEtran}
\usepackage{overpic}
\usepackage{times}
\usepackage{epsfig}
\usepackage{graphicx}
\usepackage{subfigure}
\usepackage{amsmath}
\usepackage{amssymb}
\usepackage{url}
\usepackage{graphicx}
\usepackage{animate}
\usepackage{multirow}
\usepackage{color,xcolor}
\usepackage[T1]{fontenc}

\usepackage{tikz}
\usetikzlibrary{spy}
\usepackage{xspace}
\newcommand{\etal}{{\emph{et al.}}}

\ifCLASSINFOpdf
\else
\fi

\usepackage[switch,columnwise]{lineno}

\begin{document}
	%
    \title{Face Normal Estimation from Rags to Riches}
	%
	%
	
	\author{Meng Wang, Wenjing Dai, Jiawan Zhang, and Xiaojie Guo
	\thanks{M. Wang is with the School of Computer Science and Technology, Tiangong University. Wenjing Dai is with the Algorithm Department of Fitow (Tianjin) Detection Technology Co., Ltd. J. Zhang and X. Guo are with the College of Intelligence and Computing, Tianjin University, Tianjin 300350, China.  E-mail: ({autohdr,abigail.dai4}@gmail.com, jwzhang@tju.edu.cn, xj.max.guo@gmail.com). X. Guo is corresponding author.}
	}
	
	%
	%

\markboth{Journal of \LaTeX\ Class Files,~Vol.~14, No.~8, August~2015}%
{Shell \MakeLowercase{\textit{et al.}}: Bare Demo of IEEEtran.cls for IEEE Journals}
%


    \maketitle

\begin{abstract}

Although recent approaches to face normal estimation have achieved promising results, their effectiveness heavily depends on large-scale paired data for training. This paper concentrates on relieving this requirement via developing a coarse-to-fine normal estimator. Concretely, our method first trains a neat model from a small dataset to produce coarse face normals that perform as guidance (called exemplars) for the following refinement. A self-attention mechanism is employed to capture long-range dependencies, thus remedying severe local artifacts left in estimated coarse facial normals. Then, a refinement network is customized for the sake of mapping input face images together with corresponding exemplars to fine-grained high-quality facial normals. Such a logical function split can significantly cut the requirement of massive paired data and computational resource. Extensive experiments and ablation studies are conducted to demonstrate the efficacy of our design and reveal its superiority over state-of-the-art methods in terms of both training expense as well as estimation quality. Our code and models are open-sourced at: \url{https://github.com/AutoHDR/FNR2R.git}.

\end{abstract}



%
\IEEEpeerreviewmaketitle

\section{Introduction}\label{sec:intro}

\IEEEPARstart{F}{ace} normal estimation is crucial to understanding the 3D structure of facial images, which acts as a fundamental component for various applications such as portrait editing \cite{shu2017neural,sengupta2018sfsnet,ren2021pirenderer} and augmented reality \cite{choudhary2017real,guo2023vid2avatar}. However, this problem is in nature highly ill-posed, since infinite recoveries from an image are feasible. It is difficult to determine which one is correct, without additional constraints. To address the ill-posedness, a number of approaches~\cite{shu2017neural,trigeorgis2017face,sengupta2018sfsnet,yu2021heatmap} have been presented over the past decades, which learn to recover facial components from a single image by exploring essential geometric information about human faces. 


Most, if not all, of contemporary methods rely on large-scale paired data to achieve the goal. However, on the one hand, it is difficult to collect sufficient real-world data in practice. While on the other hand, synthetic data often lacks high-frequency realistic geometric details, models trained on such data thus fail to produce high-quality results.
To alleviate the pressure from data, some schemes \cite{sengupta2018sfsnet,trigeorgis2017face,zhou2019deep} have been recently proposed . As a representative, Abrevaya \etal~\cite{abrevaya2020cross} alternatively treated this problem as a cross-domain translation task. They trained a sophisticated network structure with deactivable skip connections. Unfortunately, their performance greatly depends on the availability of substantial ground-truth data. For loosing the dependence on extensive paired data, Wang \etal~\cite{wang2022towards} developed a two-stage training network. At the first stage, they acquired knowledge of face normal priors from limited paired data, while at the second stage, they enhanced the learned normal priors by integrating them with facial images to produce high-quality face normals.

\begin{figure}[t]
	\begin{center}
		\begin{overpic}[width=\linewidth]{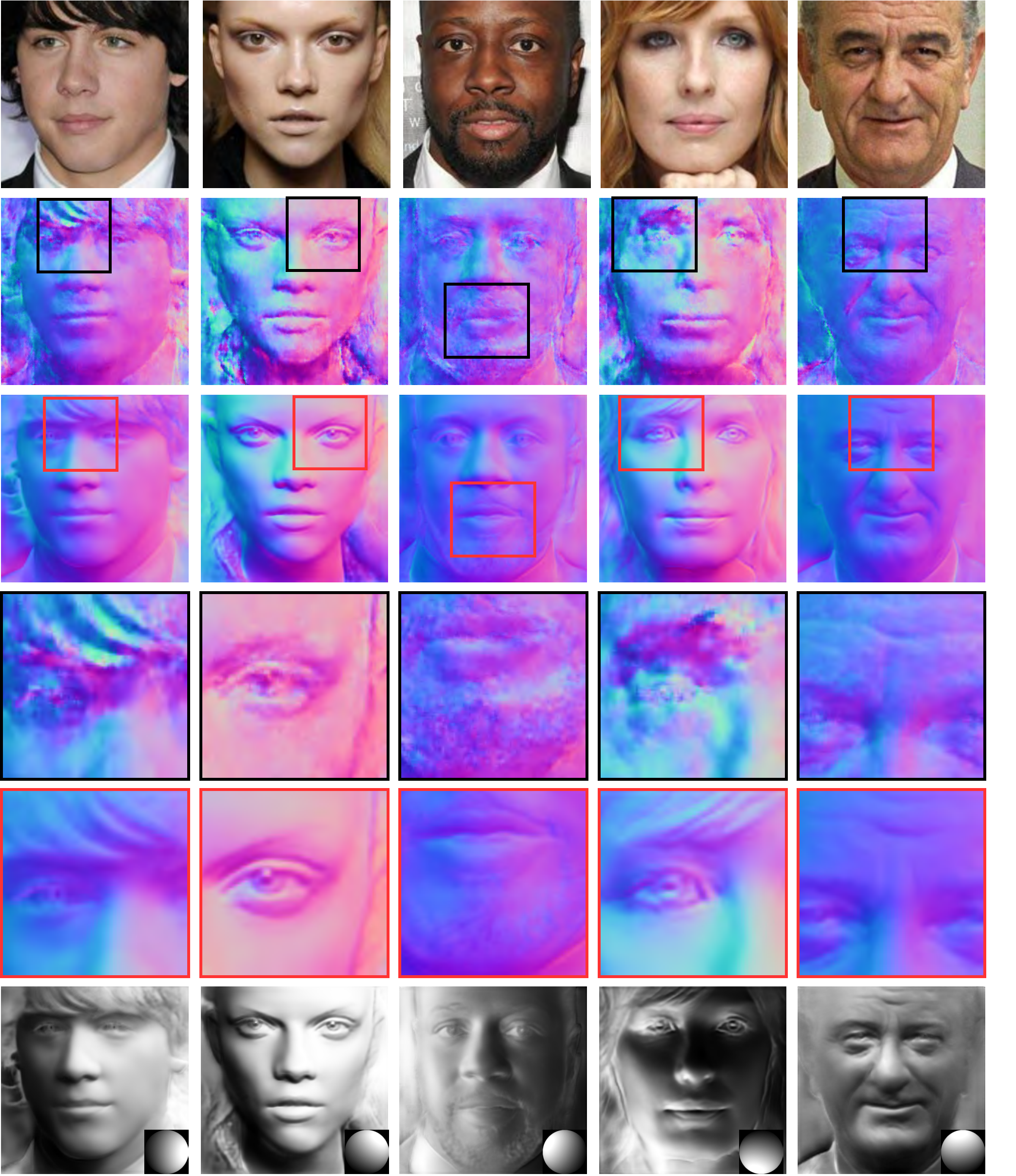}
  	\put(84.5, 94){\footnotesize{\rotatebox{270}{Input}}}
		\put(84.5, 81){\footnotesize{\rotatebox{270}{Ex-normal}}}
		\put(84.5, 64){\footnotesize{\rotatebox{270}{Re-normal}}}
        \put(84.5, 46){\footnotesize{\rotatebox{270}{Ex-zoom}}}
        \put(84.5, 30){\footnotesize{\rotatebox{270}{Re-zoom}}}
        \put(84.5, 12){\footnotesize{\rotatebox{270}{Shading}}}
		\end{overpic}
        \vspace{-6mm}
		\caption{Our model can generate high-quality face normals from single face images. The terms ``Ex-normal" and ``Re-normal" refer to the coarse exemplar normal, its refined version. ``Ex-zoom" and ``Re-zoom" denotes the zoomed-in exemplar and the zoomed-in refinement normal.}
		\label{fig:teaser}
	\end{center}
\end{figure}
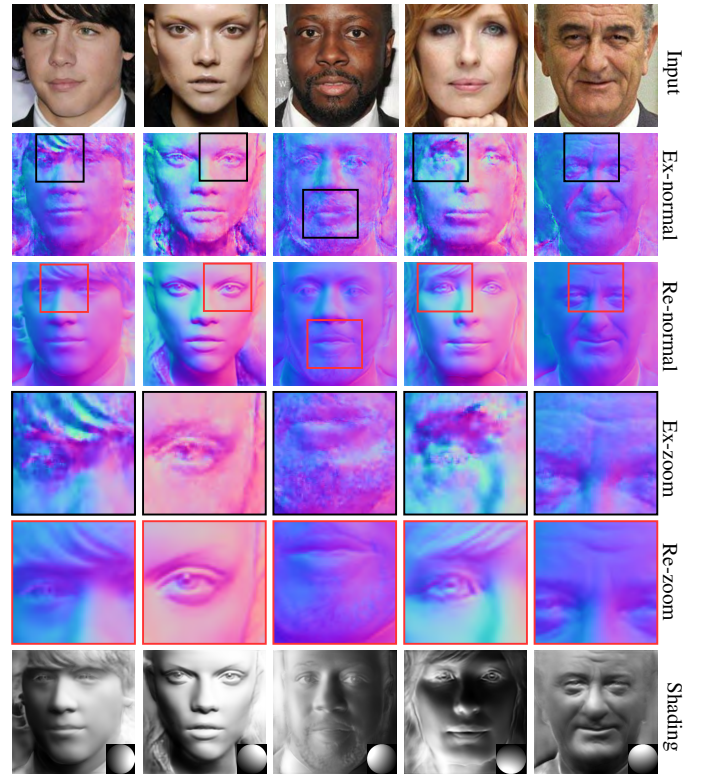

Despite remarkable progress made over last years, several challenges still remain. First, the scarcity of accurately annotated real-world data limits the performance of model. Moreover, the gap between synthetic and real data poses an obstacle, models trained on such synthetic data frequently lose their power on real-world images. Existing methods struggle with high-frequency geometric details, compromising the fidelity of estimated face normals. In addition, environmental fluctuations, like changes in background or pose, is another factor affecting the performance. The absence of multi-scale facial structural features during training further reduces geometric detail fidelity.

To tackle the aforementioned challenges, we present an innovative and effective solution that significantly reduces the need of extensive training data derived from scanned 3D models or large-scale datasets with ground truth. Our approach capitalizes on the power of exemplar-based deep learning to estimate face normals from a single image in real-world settings. To commence the process, we train a network on a modest dataset, enabling us to extract coarse normals from input facial images. By focusing on a smaller dataset, we overcome the heavy reliance on the availability of large-scale carefully-annotated data, and achieve a more practical and accessible solution. To further refine the estimation, we leverage the estimated coarse normals in an exemplar-based mapping network. This network takes both the original face image and the corresponding coarse exemplar as inputs to generate high-quality normals. 

Conceptually, our network is divided into three sub-networks, including the exemplar encoding network, the face encoding network, and the feature injection network. The exemplar encoding network generates an intermediate latent representation of exemplar. The face encoding network focuses on learning facial geometric features, while the feature injection network employs the feature modulation~\cite{karras2020analyzing} to synthesize high-fidelity normals. This is accomplished by modulating feature weights using both global and local facial geometric information for multiple iterations, resulting in final high-quality normals (please see Fig.~\ref{fig:teaser} for examples). These three sub-networks work together and are learned using simple perceptual and reconstruction losses. The major contributions of this work can be summarized as follows:
\begin{itemize}
	\item We propose to logically split the face normal estimation process into two components. One component attempts to predict coarse results, while the other performs refinement. This manner can significantly mitigate the pressure from paired data for practical use.
	
	\item We propose a novel framework for high-fidelity face normal estimation, addressing critical limitations in existing approaches regarding preserving fine-grained structural details and consistency. Furthermore, a novel normal refinement network is designed to refine coarse-normal predictions by integrating original facial features and coarse-normal outputs, ensuring high-quality and structurally consistent final results.
    
	
	
	\item Extensive experiments and ablation studies are conducted to demonstrate the effectiveness of our design, and reveal its superiority over current state-of-the-art methods.
\end{itemize} 



The previous version of this manuscript, referred to as `HFN', was published in ~\cite{wang2022towards}. Building upon the foundation laid in `HFN', this journal version takes a further step towards optimizing the model architecture by removing redundant parameters. By doing so, we can not only improve computational efficiency but also enhance the interpretability of design. In addition, we introduce a self-attention mechanism in the first stage, which enables the network to learn a normal distribution tailored to the facial structure, resulting in more accurate and robust normal estimation during the refinement stage. By attending to the most relevant facial features, the self-attention mechanism helps to capture fine-grained details and subtle variations in the facial geometry. Based on comprehensive experimental evaluation, we analyze the effectiveness of each component in our design, shedding light on their individual contributions to the overall performance. Moreover, we demonstrate the advantages of our method for face normal estimation through quantitative metrics and qualitative visual results. The experimental findings reinforce the efficacy and robustness of our proposed approach. To promote future research, encourage collaboration, and facilitate comparisons from the community, we have made the code publicly available at \url{https://github.com/AutoHDR/FNR2R.git}.

\section{Related work}
This section offers a concise overview of representative works on surface normal estimation and cross-domain tasks, which are closely related to this paper.

\subsection{Surface normal estimation}
Shape from shading (SfS) \cite{horn1975obtaining} aims to recover 3D surface information from images based on shading cues, which is typically formulated under the Lambertian model \cite{barron2011high,barron2012shape,barron2014shape,xiong2014shading,yang2018shape}. For exemplar, Barron \etal~\cite{barron2014shape} introduced an approach that incorporates various priors on shape, reflectance, and illumination to tackle the SfS problem. They developed a multi-scale optimization technique to estimate the shape of objects by considering multiple levels of detail. Xiong \etal~\cite{xiong2014shading} proposed a framework based on a quadratic representation of local shape, which is capable to capture the underlying geometric structure effectively. However, the mentioned SfS approaches rely on certain assumptions that may not hold in unconstrained environments, limiting their applicability to real-world scenarios. 

Alternatively, combining data-driven strategies with SfS regularizers for surface normal estimation \cite{shu2017neural,sengupta2018sfsnet} has shown the advance. Shu \etal~\cite{shu2017neural} proposed an end-to-end generative network that learns a face-specific disentangled representation of intrinsic face components. It attempts to capture the complex variations in facial shape by training a generative model directly on a large-scale face dataset. However, the smoothness constraint imposed in \cite{shu2017neural} often causes the loss of high-frequency information. While Sengupta \etal~\cite{sengupta2018sfsnet} introduced a two-step training strategy for surface normal estimation, leveraging both synthetic and real-world data. 
Although this approach has shown promising results, the reliance on prior knowledge gained purely from synthetic data in \cite{sengupta2018sfsnet} limits the applicability in real-world scenarios. The inherent gap between synthetic and real data leads to degraded performance in practice.



Our work is related to those deep learning based methods for face normal recovery from images, such as \cite{zhu2016face,feng2018joint,jackson2017large,tuan2017regressing,tran2018nonlinear,chinaev2018mobileface,tran2019learning}. These approaches aim to estimate the face normal as part of the overall 3D information recovery, rather than focusing solely on targeted normal estimation. For example, Tran \etal~\cite{tran2019towards} proposed a dual-pathway network that learns additional proxies to bypass strong regularization and enhance the quality of high-frequency details. 
However, the quality of high-frequency detail has a large room to improve. In contrast, our work introduces exemplar-based refinement, which leverages a coarse normal as an exemplar for normal refinement. Our method is more flexible and robust in face normal estimation, which can improve the accuracy and realism of face normals.

\begin{figure*}[t]
	\begin{center}
		\begin{overpic}[scale=0.32]{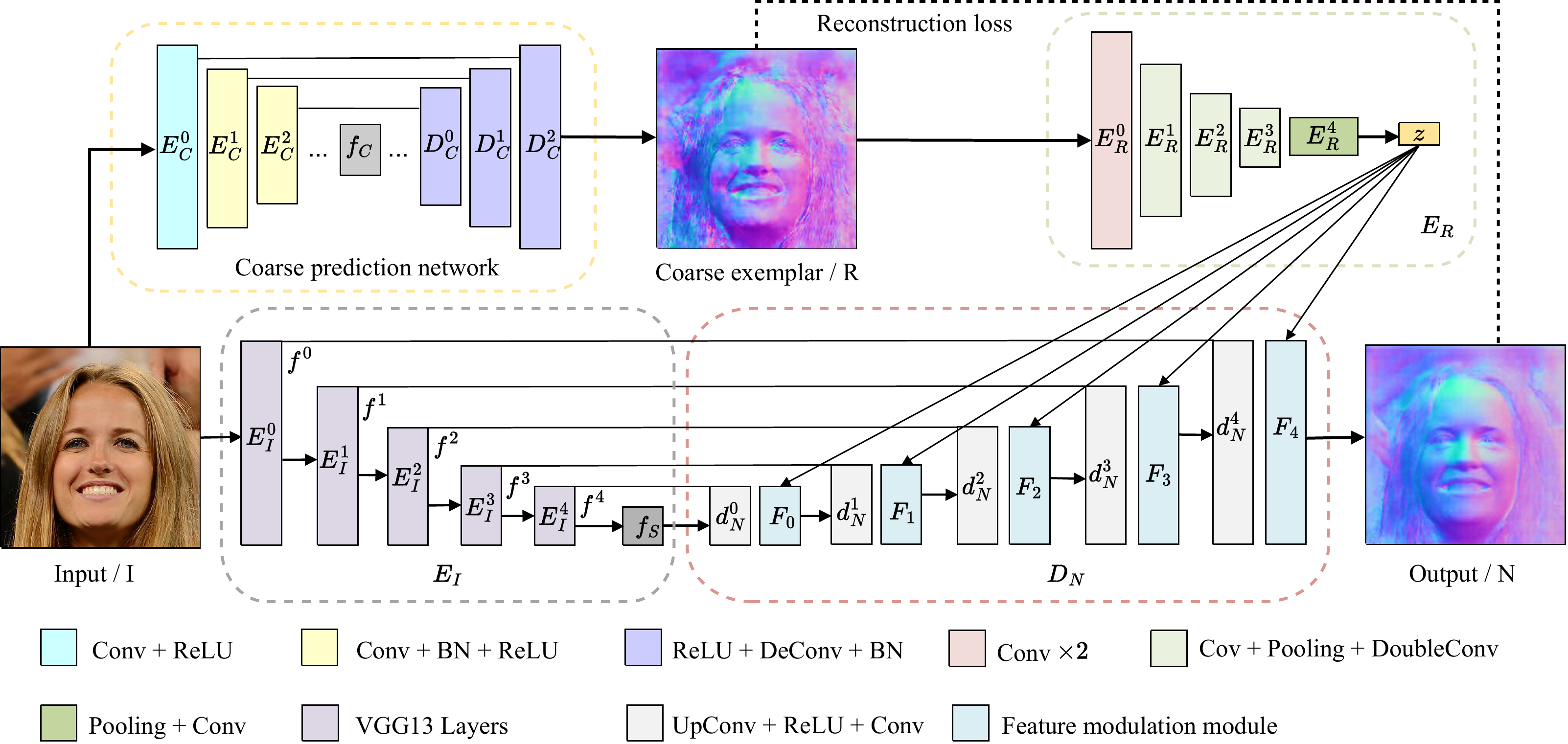}
		\end{overpic}
		\vspace{-5pt}
		\caption{Illustration of our exemplar-based face normal refinement framework that can generate high-quality normal by decoding the coarse exemplar and face structure features.}
		\label{network}
	\end{center}
\end{figure*}

\subsection{Cross-domain tasks}
The cross-modal/cross-domain learning has gained significant attention in various research communities, including style transfer \cite{gatys2016image,luan2017deep,zhang2020cross,zhan2021unbalanced,kang2021multiple}, image inpainting \cite{yu2018generative,yeh2017semantic,pathak2016context,iizuka2017globally}, and image colorization \cite{deshpande2017learning,cheng2015deep,he2018deep,zhang2019deep,xu2020stylization}. Exemplar-based learning is one such cross-domain method that utilizes both an input image and an exemplar image to generate a target image with the desired content from the input image and the style from the exemplar image. 
The existing literature on exemplar-based surface normal estimation is limited. One notable study is by Huang \etal~\cite{huang2007examplar} proposed to estimate surface normals from single images. However, this method greatly relies on a database with known 3D models.
While this approach is effective in certain scenarios, it cannot be applied on complex cases, and hardly leverages the potential of deep neural networks for capturing high-frequency information. More recently, with the emergency of deep neural networks, there have been efforts to explore exemplar-based techniques for surface normal estimation. Abrevaya \etal~\cite{abrevaya2020cross} introduced a cross-modal approach for synthesizing face normals, which enables the exchange of facial features between the image and normal domains through deactivated skip connections. But, both the methods proposed in \cite{huang2007examplar} and \cite{abrevaya2020cross} suffer from the loss of high-frequency details in estimated normals. It is worth to emphasize that, our exemplar-based face normal refinement aims to overcome the limitations by reducing the need for extensive data collection, and achieve high-quality estimation of surface normals more efficiently and effectively.

\section{Method}
Our goal is to estimate high-quality face normals from single images in a two-stage fashion. An the first stage, we train the coarse normal predictor \emph{CP-Net} using the Photoface dataset \cite{zafeiriou2011photoface}, which provides ground truth for face normals. However, there is a domain gap between the training data and real-world face images captured in diverse natural scenes. To narrow the gap between the coarse estimation and the desired fine-grained output, we further propose a normal refinement network (\emph{NR-Net}) as the second stage. In our proposed framework, the predicted coarse normal from the first stage serves as an exemplar, which is combined with the input face image for the second stage of normal refinement.
In subsequent sections, we shall provide more detailed information about the design and implementation.

\subsection{Overview}

The whole framework is schematically shown in Fig. \ref{network}, The first stage trains the \emph{CP-Net} to produce a coarse exemplar (as reference $R$) simply using the generator architecture proposed in \cite{isola2017image}. This network contains a relatively small number of parameters, which enables fast training. To ensure the accuracy of the generated normal distribution for corresponding parts of face, we introduce a self-attention-based discriminator. This discriminator provides additional constraints during the training process, guiding the \emph{CP-Net} to produce the consistency of the generated normal distribution. In the second stage, we construct a normal refinement network (\emph{NR-Net}) that fully leverages both face structure features $f^{4}$ and normal features $\mathbf{z}$. This structure allows us to effectively combine the coarse exemplar $R$ with the detailed information present in the original face image $I$. To facilitate features fusion, we introduce a feature modulation module~\cite{karras2020analyzing}d that effectively integrates the face structure features and normal features to produce high-fidelity face normal $N$.


Specifically, the coarse exemplar $R$ is produced from an input face image $I$ by the \emph{CP-Net} as $R = {\emph{CP-Net}}(I)$.
The coarse exemplar $R$ is encoded by the normal feature encoder $E_{R}$ to form normal features $\mathbf{z} = E_{R}(R)$.
Additionally, the face structure encoder $E_{I}$ can extract facial structure features $\mathbf{f_{l}}$ as follows:
\begin{equation}
\mathbf{f}_{i}=\left\{
\begin{array}{l}
		E_{I}^{i}(I),            \text { if } i=0,\\
		E_{I}^{i}(f^{i-1}),      \text { otherwise},
\end{array}\right.	
	\label{FaceEncoder}
\end{equation}
where $i$ ranges from $0$ to $5$, standing for the layer index of face structure encoder $E_{I}$. In addition, $f^{5}=f^{s}$ in Fig \ref{network}. 
The normal refinement network, denoted as $D_{N}$, takes the modulated features and face structure features as input and produce the refinement high-fidelity normal, which  can be described as follows:
\begin{equation}
	N = D_{N}(d_{N}^{j}, F_{j}).
	\label{decoderNormal}
\end{equation}
 $N$ is the final normal and $d_{N}^{j} (j\in [0,4])$ represents the face structure features obtained via a deconvolution layer followed by two convolution layers. $F_{j}$ corresponds to the modulated features generated by the feature modulation module.

\begin{figure}[t]
	\begin{center}
		\begin{overpic}[scale=0.3]{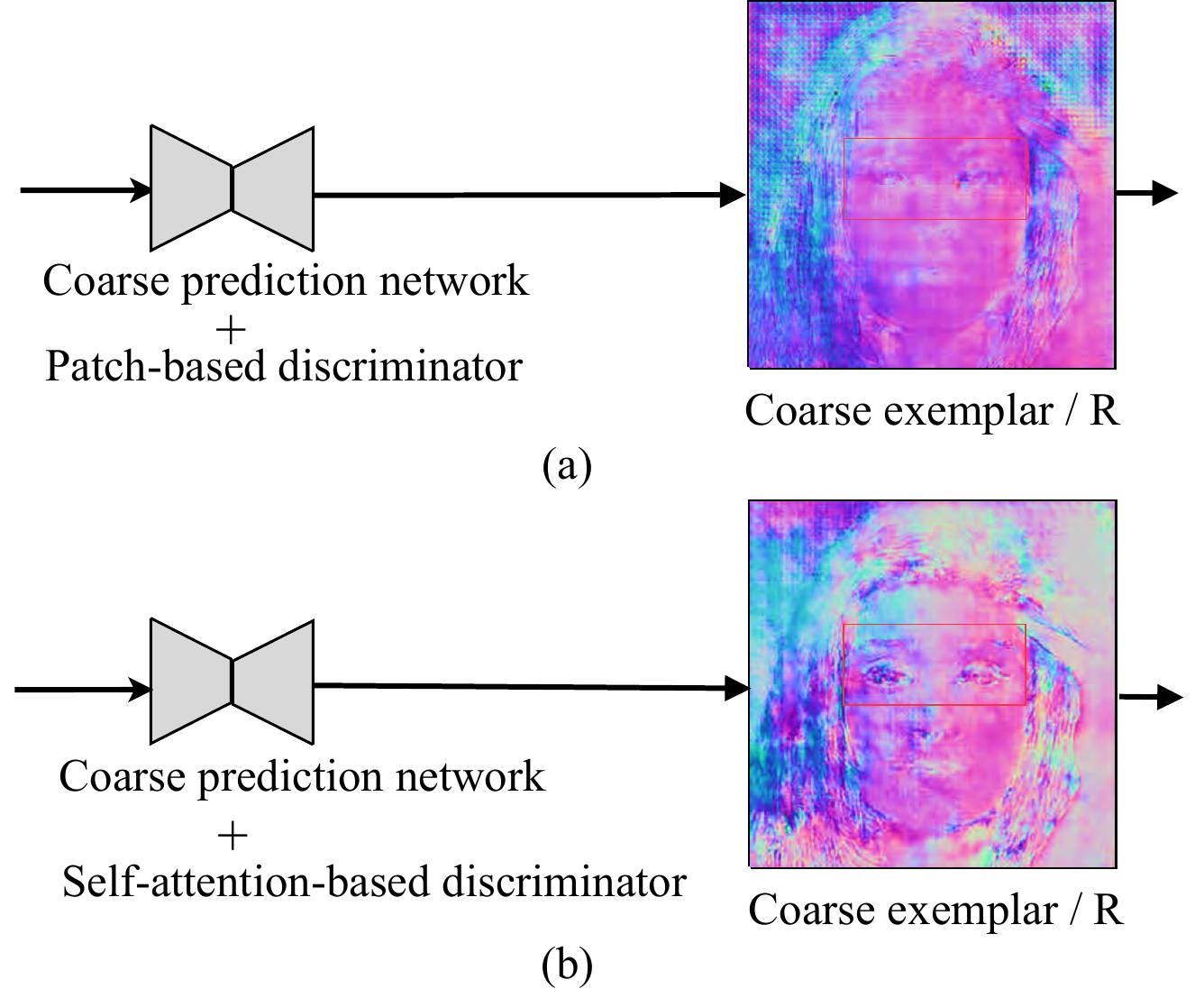}
		\end{overpic}
		\vspace{-13pt}
		\caption{The coarse prediction network \emph{CP-Net} comparison between ours (b) and `HFN'~\cite{wang2022towards} (a). Note that our coarse examplar/R is able to capture the correlation of normal direction changes where the structure changes greatly.}
		\label{fig:pretrainGAN_COM}
	\end{center}
\end{figure}

\subsection{Feature modulation module} 
Based on previous studies~\cite{karras2020analyzing,jang2021stylecarigan,zhao2021color2embed}, the feature modulation allows for trainable determination of feature weights. In order to leverage both the structural features that capture fine face details, and the normal features that represent the normal distribution, we incorporate the feature modulation module~\cite{karras2020analyzing} into our normal refinement network. This module dynamically adjusts the weights of normal features based on structure features, employing a multi-scale injection in the feature space. Mathematically, it can be expressed as follows:
\begin{equation}
	\mathbf{\bar{w}} = \mathbf{w} \cdot \mathbf{s} \cdot {\text {Linear}}(\mathbf{z}),
	\label{InjectWeigh}
\end{equation}
where $\mathbf{w}\in \mathbb{R}^{C_{i} \times C_{j} \times K \times K}$ and $\mathbf{\bar{w}}\in \mathbb{R}^{C_{i} \times C_{j} \times K \times K}$ are the original weights and modulated convolution weights, respectively. $K$ is the kernel size, $C_{i}$ and $C_{j}$ denote the numbers of input and output channels, respectively. The function $\text{Linear}(\cdot)$ performs a mapping operation. It takes the normal features $\mathbf{z^{i}}$ as input and maps them to the weights at the feature scale $\mathbf{s}$ that corresponds to the feature map. This mapping function allows the normal features to influence the weights of the convolutional layers, enabling the modulation of the network's behavior based on the specific normal distribution. The normalization can be written as:
\begin{equation}
	\text {Norm}(\mathbf{\bar{w}})=\frac{\mathbf{\bar{w}}}{\sqrt{\sum \mathbf{\bar{w}}^{ 2}+\epsilon}},
	\label{FeatNormalization}
\end{equation}
where $\epsilon$ is a small positive constant used to avoid division by zero. To ensure the outputs normalized with a unit standard deviation, we further normalize the dimension of $\mathbf{\bar{w}}$. Given the facial structure features $\mathbf{f}_{j}$, the modulated features $\mathbf{m}$ are obtained via:
\begin{equation}
	F_{j}=\text {Conv}\left(\text{Norm}(\mathbf{\bar{w}}), \mathbf{f}_{j}\right),
	\label{ModuWeight}
\end{equation}
where $\text{Conv}(\cdot)$ represents a convolution operation. 

\begin{figure}[t]
	\begin{center}
		\begin{overpic}[width=\linewidth]{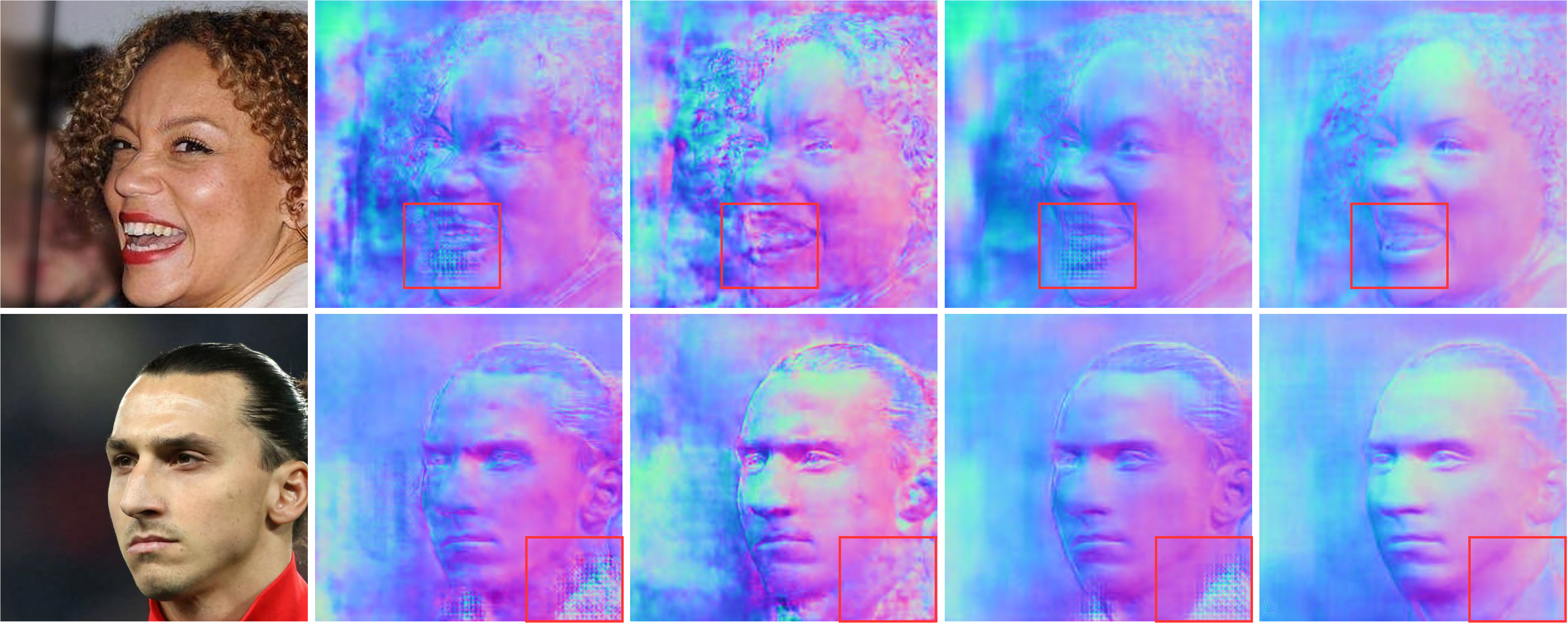}
			\put(5, -3.5){\footnotesize{Input}}
			\put(22.5, -3.5){\footnotesize{HFN-C}}
			\put(44, -3.5){\footnotesize{Ours-C}}
			\put(64, -3.5){\footnotesize{HFN-R}}
			\put(85, -3.5){\footnotesize{Ours-R}}
		\end{overpic}
		\vspace{-10pt}
		\caption{The comparison between coarse and refined normal estimations on the FFHQ dataset~\cite{karras2019style}. The labeled as `HFN-' represent the outputs generated by \cite{wang2022towards}. The labels `-C' and `-R' indicate the coarse normal and refinement normal, respectively.}
		\label{self_explain}
	\end{center}
\end{figure}

\begin{figure*}[t]
	\begin{center}
		\begin{overpic}[scale=0.37]{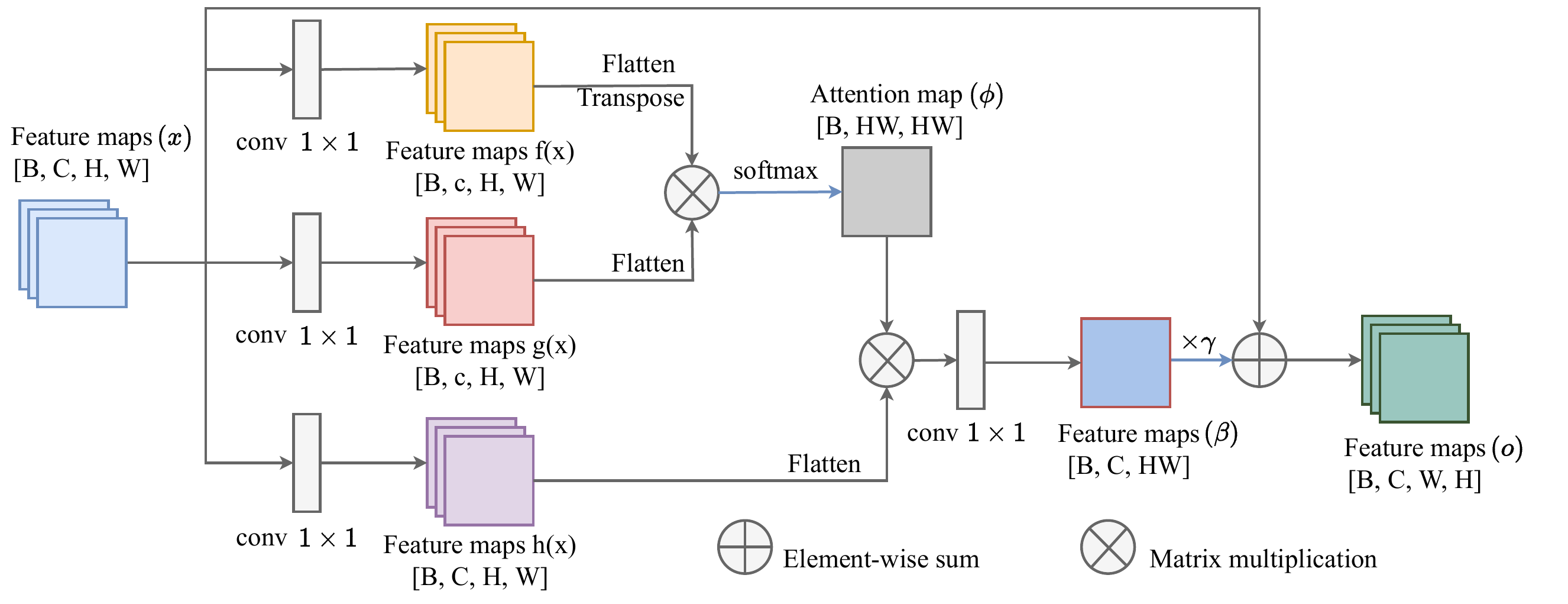}
		\end{overpic}
  	\vspace{-15pt}
		\caption{The self-attention module in our framework operates on feature maps $x$ with dimensions $B \times C \times H \times W$, where $B$ denotes the batch size, $C$ represents the number of channels, and $H$ and $W$ denote the height and width, respectively. In the self-attention module, we set the intermediate channel size $c$ as $C/8$, and $\gamma$ is a scalar parameter that is learned during training.}
		\label{fig:attention}
	\end{center}
\end{figure*}

\subsection{Self-attention module}
In `HFN' \cite{wang2022towards}, the generated normals exhibit apparent artifacts (see Fig. \ref{self_explain} in the mouth and neck areas for examples). To address this issue, it is important to extract features in a manner that incorporates attention-driven mechanisms and considers long-range dependencies. Taking inspiration from the self-attention mechanism \cite{vaswani2017attention,wang2018non,xu2020mef}, we introduce a self-attention module in the discriminator for our first stage training. For instance, as illustrated in Fig.~\ref{fig:pretrainGAN_COM}, we can acquire a more comprehensive understanding of the structural characteristics compared to `HFN'. This module enables our model to learn long-range dependencies and improve the overall performance during second stage refinement. As shown in Fig \ref{self_explain}, our approach successfully preserves facial structure details.

In Fig. \ref{fig:attention}, we present the architecture of the self-attention module, which consists of three standard $1\times1$ convolution layers.
The feature maps $f(x)$ are first flattened and transposed, while $g(x)$ is only flattened. These two sets of feature serve as the query and key-value pairs for computing attention maps. To obtain the attention map $\phi$, the flattened and transposed $f(x)$ is multiplied with $g(x)$ and then processed by the softmax function.
Similarly, 
the feature map $h(x)$ is also flattened and multiplied with the attention map $\phi$ to produce the feature maps $\beta$. The final feature maps $o$ are obtained by combining the feature maps $\beta$ with the input feature maps $x$ using the formula $o = \gamma \times \beta + x$, where $\gamma$ is a learned parameter. This combination of features allows the network to enhance relevant information based on the attention map and preserve important details from the input. 

\subsection{Normal feature encoder}

\begin{figure}[t]
	\begin{center}
		\begin{overpic}[scale=0.29]{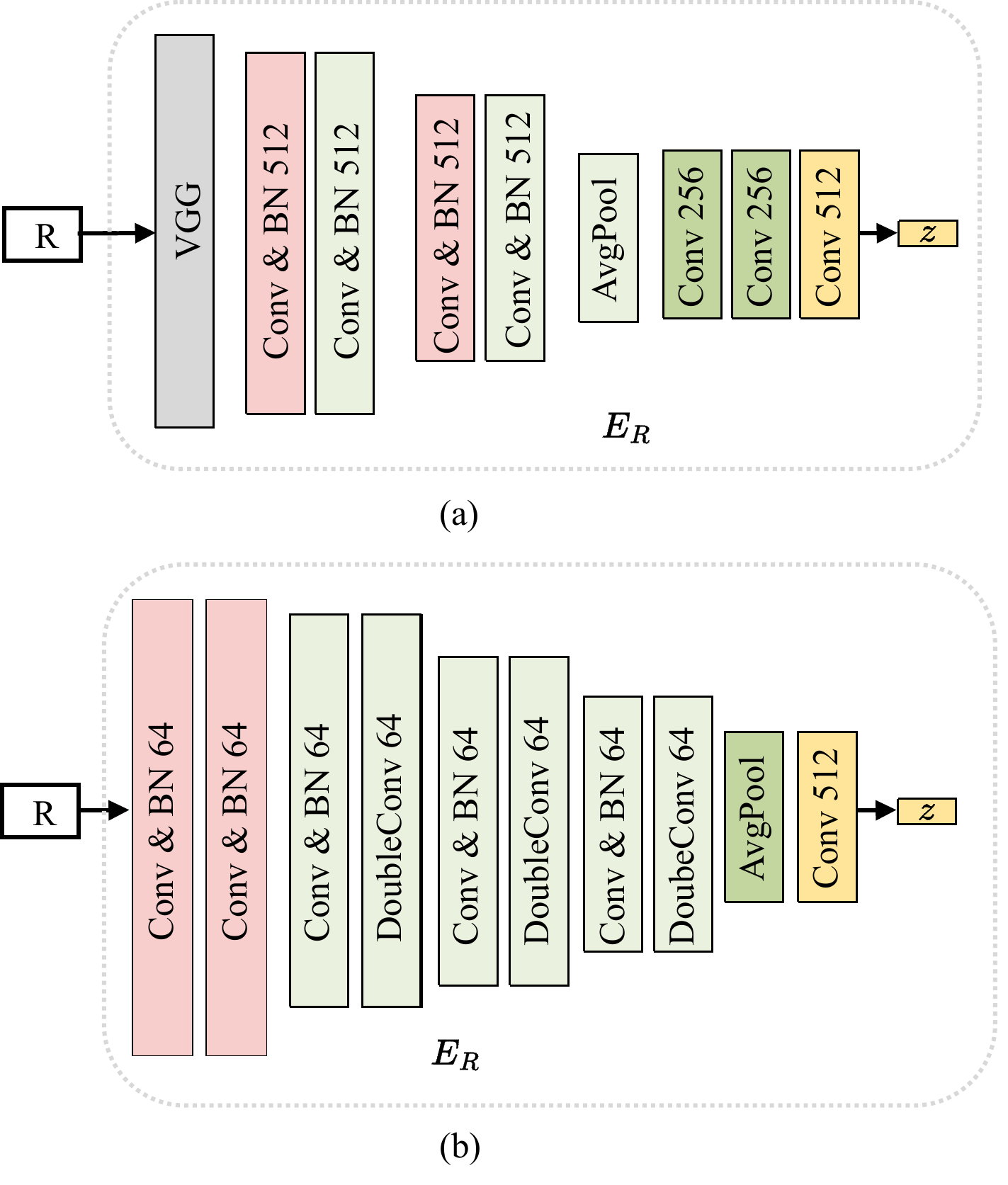}
		\end{overpic}
		\vspace{-15pt}
		\caption{Normal features encoder network ($E_{R}$) comparison between ours (b) and `HFN'~\cite{wang2022towards} (a). `R' represents as coarse examplar/R produced from the coarse prediction model.}
		\label{fig:NormalEncoderNet}
	\end{center}
\end{figure}

VGG19 is renowned for its capability to learn intricate patterns via a multilayer nonlinear structure. However, the normal feature space has relatively low dimensionality (a 256-dimensional vector in this paper), which results in an excessive amount of parameter redundancy. This redundancy causes unnecessary computational overhead and is unsuitable for efficiently training our refinement network. Thus, we design a smaller network based on the Feature Pyramid Network (FPN) structure \cite{lin2017feature} specifically tailored for normal distribution learning. On the other hand, the FPN-based network has multiple convolution layers before pooling, enabling it to construct stronger feature representations while retaining spatial information. 
To highlight the difference between our network architecture and `HFN' \cite{wang2022towards}, we provide a comparison in Fig.~\ref{fig:NormalEncoderNet}.
Compared to Fig.~\ref{fig:NormalEncoderNet} (a), our normal feature encoder $E_{R}$ employs a three-level FPN structure \cite{lin2017feature}. The layers preceding the adaptive average pooling are responsible for reducing spatial dimensions and extracting normal features. Subsequently, an adaptive average pooling layer is applied to aggregate these features, followed by a convolution layer. The output is denoted as $\mathbf{z}$ and represents the final normal distribution feature utilized for normal refinement.

\subsection{Architecture}

\textbf{The architecture of \emph{CP-Net}.} 
In the refinement stage, the exemplar plays a crucial role in determining the normal distribution. To effectively learn a robust normal distribution on faces, we introduce a self-attention-based discriminator~\cite{wang2018non} to capture long-range dependencies and improve the modeling of normal distribution. The coarse prediction network \emph{CP-Net} is depicted in Fig. \ref{fig:pretrainGAN_COM}. This network architecture is based on the generative networks framework proposed in \cite{isola2017image}, which has shown promising results in image-to-image translation tasks. By leveraging the attention, our coarse exemplar is capable of learning the correlated normal distribution on different facial regions, such as around the mouth and eyes, as given in Fig. \ref{self_explain}. This modification can achieve more robust and accurate representations of normal distribution, compared to our previous `HFN'~\cite{wang2022towards}.

\textbf{The architecture of \emph{NR-Net}.} 
To address noticeable artifacts left in predicted coarse normals and generate high-fidelity face normals, the refinement network of `HFN'~\cite{wang2022towards} utilizes a VGG19~\cite{simonyan2014very} architecture to extract normal features, as shown in Fig.~\ref{fig:NormalEncoderNet} (a).  Although `HFN' is capable of removing artifacts with the assistance of the coarse exemplar, the refined normals may still exhibit artifacts in some cases, as depicted in Fig.~\ref{Com_C_H_normal}. Furthermore, it is worth noting that the size of the normal distribution feature is relatively small, say $256 \times 1 \times 1$. This indicates that the features can be effectively learned by a smaller network without compromising the quality of the generated normals. As a result, we remove the VGG network during the refinement stage, mitigating the need for excessive computational resources. Moreover, we apply a multi-scale network to learn the normal distribution across different scales of face. This allows us to capture the nuances and variations in the normal direction more effectively, ensuring the generated normals accurately represent the underlying facial structure.

\subsection{Loss function}
\textbf{The loss function of \emph{CP-Net}.}
We employ a reconstruction loss to generate coarse normal maps as:
\begin{equation}
	\mathcal{L}_{\text {Normal }}=\frac{1}{HW} \sum_i^{HW}\left(1-{N}_{gt}^{\top} \tilde{{N}}_c\right),
	\label{pre_Nloss}
\end{equation}
where $H$ and $W$ are the height and width of the images. ${N}_{gt}$ and $\tilde{{N}}_{c}$ represent the ground-truth normal and predicted coarse normal, respectively. $T$ is the transpose of normal map.
To enhance the generalization of the coarse prediction model, we incorporate an adversarial loss through a self-attention-based network, which serves as the discriminative network. The adversarial loss $\mathcal{L}_{Dcp}$ for the \emph{CP-Net} is defined as the output of the discriminator when given the predicted coarse normals $N_{c}$. The goal is to minimize the error between the predicted normal distribution and the real normal distribution. The complete loss function for \emph{CP-Net} is formulated as: 
\begin{equation}
	\mathcal{L}_{pre} = \mathcal{L}_{\text {Normal }} + \lambda_{Dcp} \mathcal{L}_{Dcp},	
	\label{pre_wholeLoss}
\end{equation}
where $\lambda_{Dcp}$ is a weighting factor that balances the importance of the normal reconstruction loss and the discriminator loss. In this case, $\lambda_{Dcp}$ is set to $0.0001$ during training.

\textbf{The loss function of \emph{NR-Net}.} 
During the refinement stage, we only adopt a normal reconstruction loss to achieve our objective. This loss has two purposes: firstly, it encourages the fine-grained output normal distribution to resemble the coarse exemplar normal; secondly, it ensures that the refined normals have a similar structure to the exemplar. The reconstruction loss function is identical to Eq. \eqref{pre_Nloss}.

\subsection{Discussion} 
\begin{figure*}[t]
	\begin{center}
		\begin{overpic}[width=\linewidth]{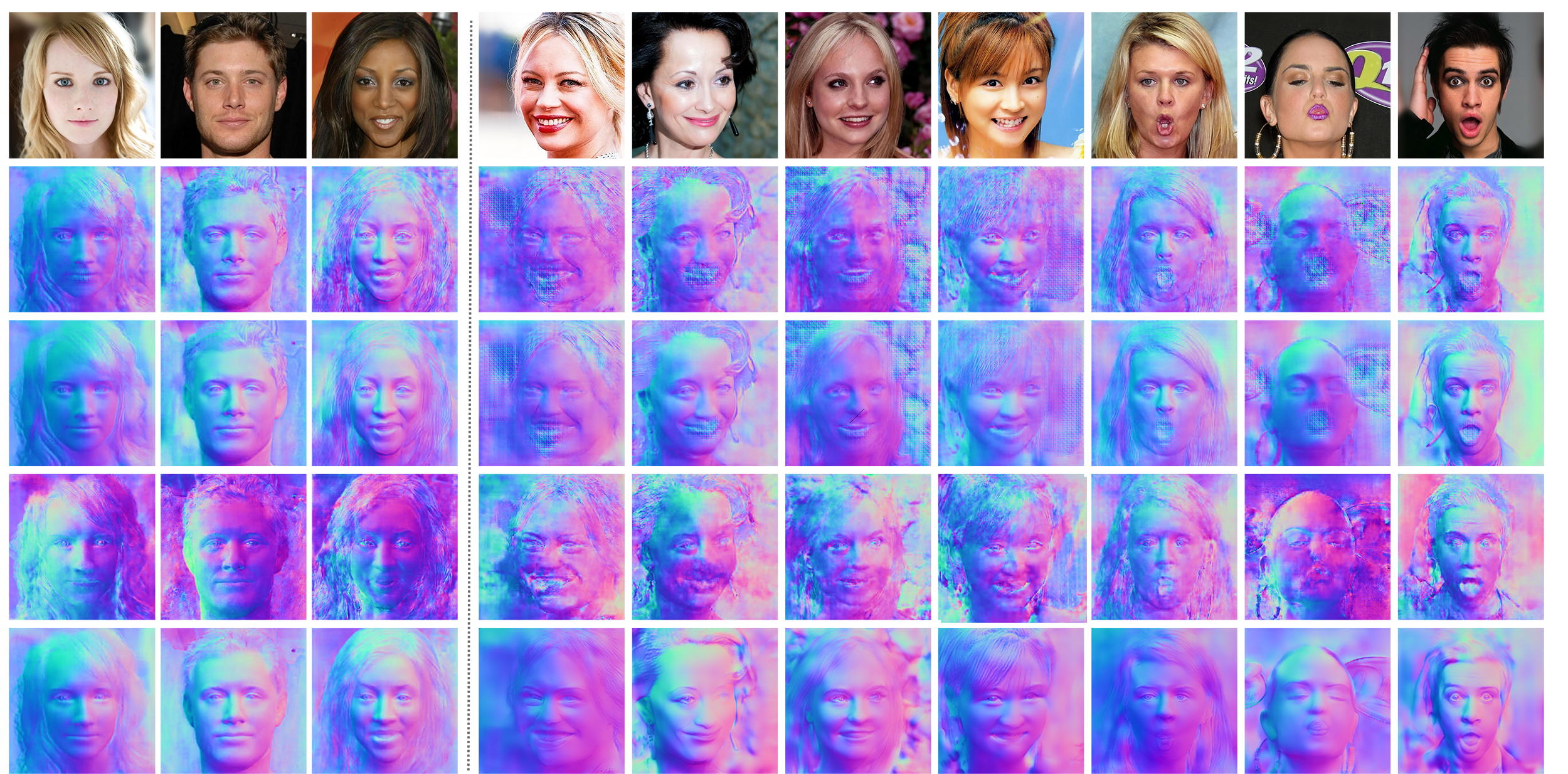}
	 \put(99, 46){\footnotesize{\rotatebox{270}{Input}}}
	 \put(99, 36){\footnotesize{\rotatebox{270}{HFN-C}}}
	 \put(99, 27){\footnotesize{\rotatebox{270}{HFN-R}}}
	 \put(99, 17){\footnotesize{\rotatebox{270}{Ours-C}}}
	 \put(99, 7){\footnotesize{\rotatebox{270}{Ours-R}}}
		\end{overpic}
		\vspace{-19pt}
		\caption{Comparison of `HFN'~\cite{wang2022towards} with our method on CelebA-HQ~\cite{karras2017progressive} is presented using coarse exemplar (labeled with `-C' ) and refined normal (labeled with `-R'). Our method outperforms `HFN' in terms of generalization ability beyond the dotted line on the right, while both methods achieve similar high-fidelity results on the left side.}
		\label{Com_C_H_normal}
	\end{center}
\end{figure*}
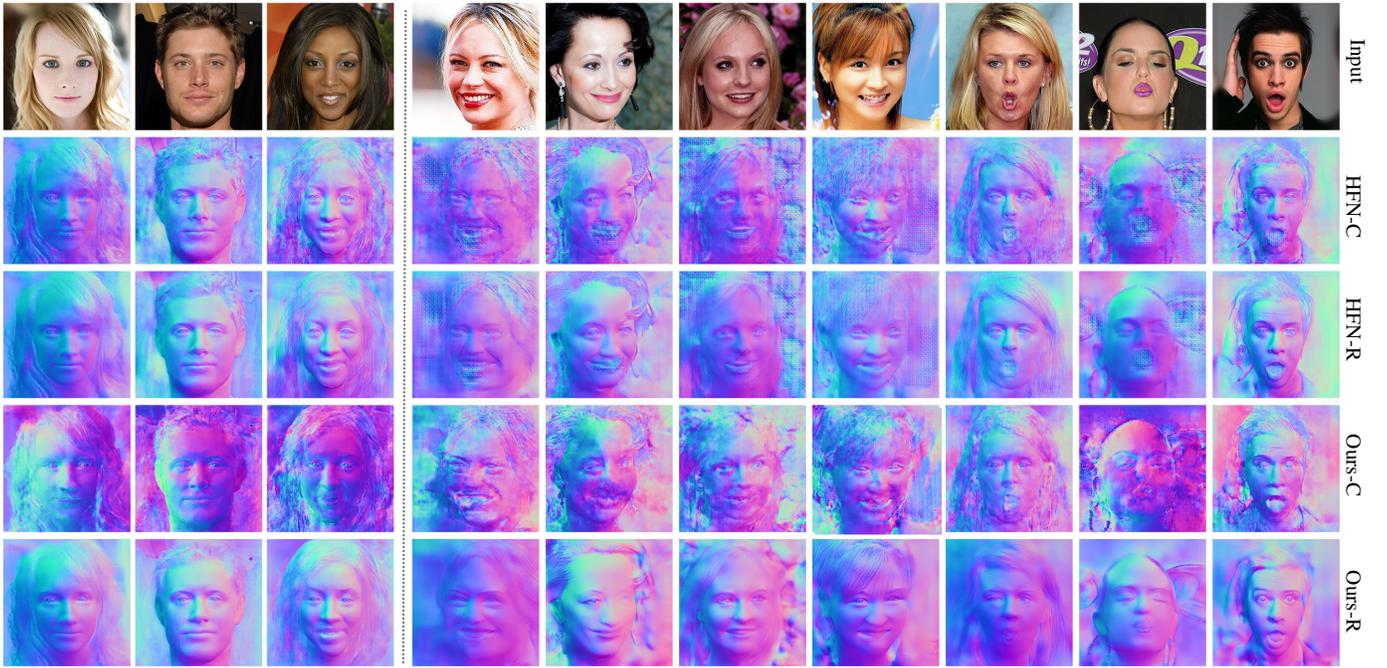

\textbf{Joint training of two stages.} 
The reason for our normal refinement framework in two steps, with a coarse prediction network \emph{CP-Net} followed by a fine-grained refinement network \emph{NR-Net}, is to address the limitations of training solely on ground truth normals. When training a model using ground-truth normals in an end-to-end manner, it heavily relies on the distribution of the training dataset. This often results in poor generalization ability when applied to real-world scenarios with diverse face poses, expressions, and backgrounds. As shown in Fig.\ref{Com_C_H_normal} and Fig.\ref{ablationNet}, the model trained on a limited dataset of facial images with ground-truth normals may exhibit poor generalization to out-of-distribution samples, leading to inaccurate predictions.
By employing a two-stage training framework, we aim to overcome the above limitations. In the first phase, a coarse normal is generated as an exemplar, providing a rough estimation of the face structure and normal distribution. This coarse normal serves as guidance for the subsequent fine-grained refinement stage. In the second step, the fine-grained normal is constructed with guidance from the exemplar, enabling the model to focus on capturing intricate normal features and facial structure features, and enabling accurate refinement of normal. 
The clear division between the two stages ensures that each phase is assigned distinct responsibilities, leading to improved performance and generalization ability. Overall, the two-stage training model can better handle diverse face images with different poses, expressions, and backgrounds. By incorporating a coarse prediction stage followed by a fine-grained refinement stage, the model can leverage the benefits of both two stages and produce high-quality normal maps that are more robust and accurate in various real-world scenarios.

\textbf{Feature injection with relatively low dimensions.} 
The use of low-dimensional features for encoding the coarse exemplar normal in our framework provides several advantages. Firstly, as depicted in Fig.~\ref{network}, during the training process, we extract low-dimensional normal features that capture the normal distribution feature of the input face. These features are not necessarily aligned with the structure but contain essential information about the face normal. Secondly, by employing low-dimensional features for the coarse exemplar, the model can avoid simply copying the coarse normal to its output during the fine-grained normal generation process. This is important because if the model were to directly copy the coarse normal, it would incur a reconstruction loss during training \cite{wang2022towards}. By utilizing low-dimensional normal distribution features as guidance, the model is trained to generate the fine-grained normal by leveraging the high-fidelity details present in the input face image. This approach ensures that the generated fine-grained normal incorporates the necessary details from the input face, as well as the accurate normal distribution on the face derived from the exemplar. The exemplar can serve as a reference for the model, guiding it to capture the essential features and accurately refine the normal distribution to achieve high-quality results.



\section{Experiments}
\subsection{Experimental Setting}
\textbf{Datasets.} 
We test our model on six different face datasets: 300-W~\cite{sagonas2013300}, CelebA~\cite{liu2015deep}, FFHQ~\cite{karras2019style}, Photoface~\cite{zafeiriou2011photoface} and Florence~\cite{bagdanov2011florence}. The 300-W dataset consists of 300 indoor and 300 outdoor images captured in real-world conditions. The CelebA is a large-scale dataset of real-world faces collected from the internet, while the FFHQ contains a diverse range of images with variations in age, ethnicity, and background. The Photoface dataset comprises photos with four distinct lighting conditions, and the ground truth normals are estimated using photometric stereo. For the Photoface dataset, we divided it into two sets: a training and a test set. The training set comprises 80\% of the data while the remaining 20\%, which consists of approximately 2.5k pairs of image/normal pairs, is used for evaluation. The Florence dataset comprises 53 3D face models. We generate facial images and their corresponding ground truth normal map to evaluate the generalization ability of our method on completely novel data.

\textbf{Metrics.} 
One metric, by following previous works~\cite{trigeorgis2017face,sengupta2018sfsnet,abrevaya2020cross}, is the mean angular error between the predicted normal and ground truth normal. To provides a more detailed evaluation of the accuracy of the estimated normal maps, we measure the percentage of pixels within the facial region with angular errors less than $20^{\circ}$, $25^{\circ}$ and $30^{\circ}$. In addition, geometric shading and normal error maps are used for qualitative comparisons, enabling a more comprehensive evaluation.

\textbf{Implementation details.}
Our framework is implemented in PyTorch~\cite{paszke2019pytorch} and was trained on a single NVIDIA 2080 Ti GPU. During training, we utilized a learning rate of $10^{-4}$ and adopted the default parameters of the Adam optimizer~\cite{kingma2020method}. Notably, we employed the discriminator~\cite{zhang2019self} as our self-attention-based discriminator. To make a fair comparison with the previous work `Cross Modal'~\cite{abrevaya2020cross} and `HFN' \cite{wang2022towards}, we follow the same approach for cropping the face images. To assess the generalization capability of our approach, we utilized a model trained on the CelebA-HQ dataset~\cite{karras2017progressive} without any face cropping, this enabled us to evaluate our model on complex scenarios. 
The \emph{CP-Net} and \emph{NR-Net} are trained for 200k iterations and 150k iterations, respectively. In the previous version, the $E_{R}$ module employed a pretrained VGG19~\cite{simonyan2014very} network to extract deep convolutional features from the `relu5\_2' layer. While in this version, we replace the original VGG19-based architecture with a more efficient and effective design, as shown in Fig.~\ref{fig:NormalEncoderNet}. This modification helps to streamline the model and improve the performance.

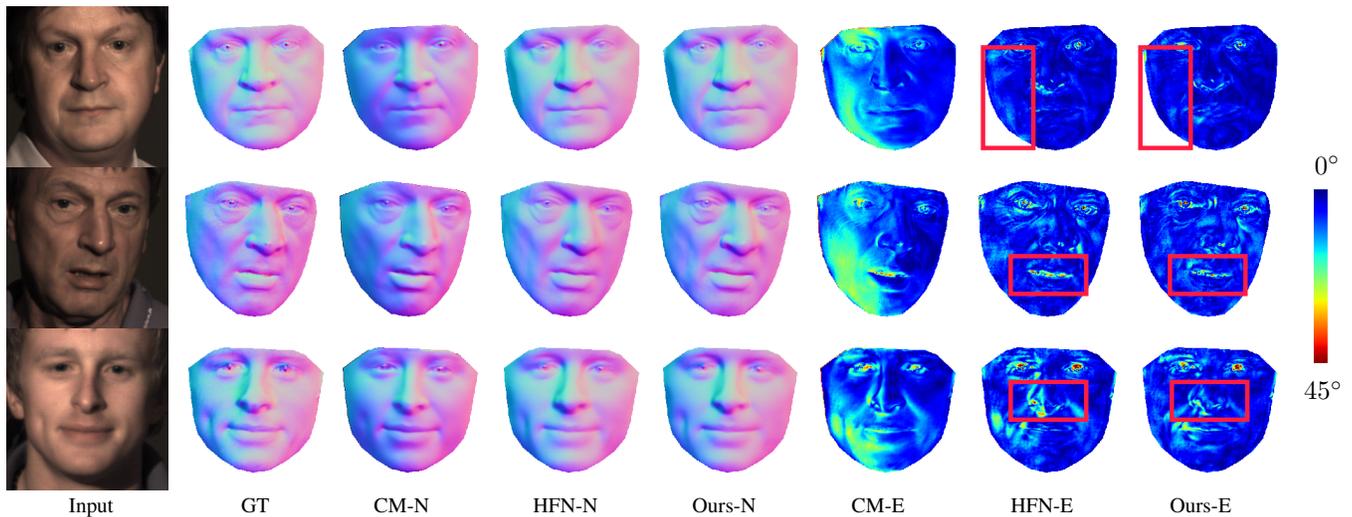
\begin{figure*}[t]
	\begin{center}
		\begin{overpic}[scale=0.33]{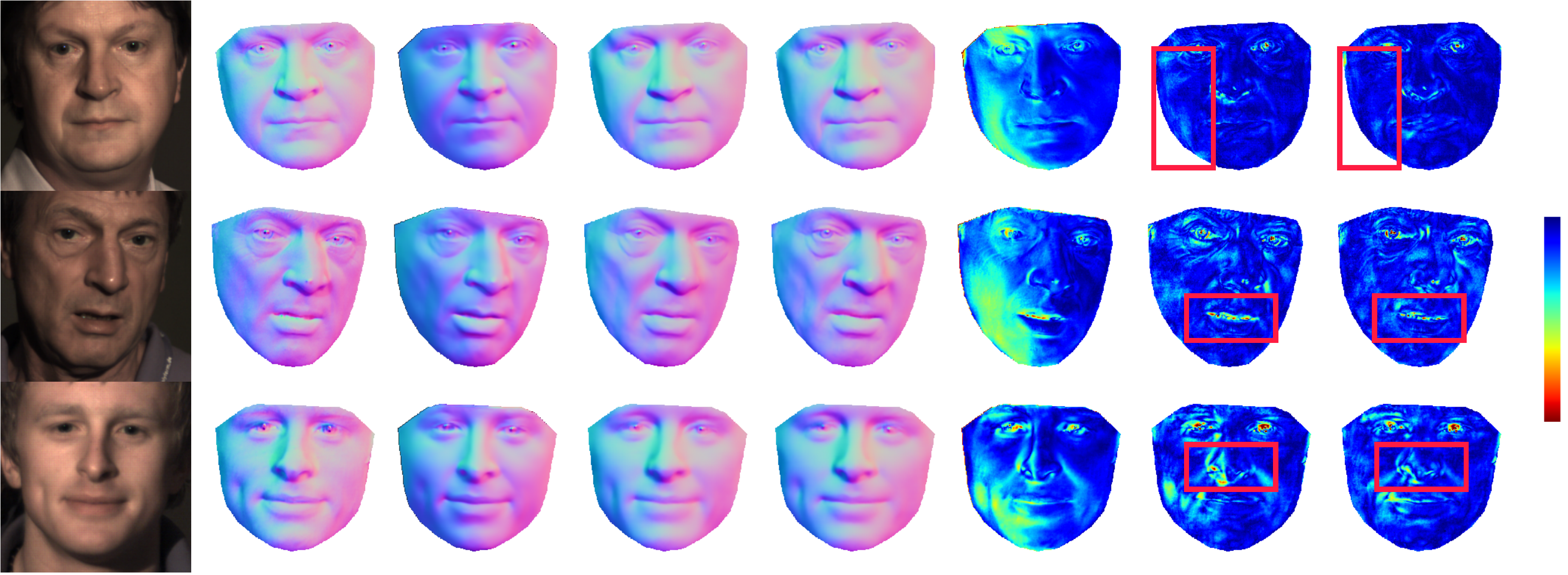}
			\put(4, -1.5){ \footnotesize{Input}}
			\put(17, -1.5){  \footnotesize{GT}}
			\put(27, -1.5){  \footnotesize{CM-N}}
			\put(39, -1.5){  \footnotesize{HFN-N}}
			\put(51, -1.5){  \footnotesize{Ours-N}}
			\put(63, -1.5){  \footnotesize{CM-E}}
			\put(75, -1.5){  \footnotesize{HFN-E}}
			\put(87, -1.5){  \footnotesize{Ours-E}}
			\put(97.9, 24){  {$0^{\circ}$}}
			\put(97.1, 7){  {$45^{\circ}$}}
		\end{overpic}
		\caption{Normal error map comparisons on the Photoface dataset~\cite{zafeiriou2011photoface}. `GT', `-N' and `-E' are the ground truth normals, predicted normals and error maps, respectively. `CM-' and `HFN-' are `Cross-modal'~\cite{abrevaya2020cross} and `HFN'~\cite{wang2022towards}. All normal error maps have an angle range of $0^{\circ}$ to $45^{\circ}$.}
		\label{F_errorMap}
	\end{center}
\end{figure*}

\subsection{Comparison}
In Fig.~\ref{Com_C_H_normal}, we can observe the comparison between the coarse exemplars (e.g. `HFN-C'~\cite{wang2022towards}/`Ours-C') and the refined high-fidelity normals (e.g. `HFN-R'/`Ours-R'). The coarse prediction model, although it converges on the training dataset and becomes increasingly similar to it, suffers from poor generalization ability due to the significant differences between the distributions of the training dataset and real-world data. While `HFN-R' can achieve more precise outcomes through secondary optimization of the normal map with the coarse normal map and input image, it may not perform optimally in certain specific areas (such as the left side of the dashed line representing the mouth in Fig.~\ref{Com_C_H_normal}). In contrast, our method is able to overcome these limitations and effectively improve the quality of the coarse exemplars, resulting in high-fidelity normals. By leveraging the guidance provided by the input face image and effectively learning from limited training data, our method is able to eliminate the artifacts and produce more accurate and visually pleasing results. 

\begin{table}[t]
	\caption{Normal reconstruction errors on the Photoface~\cite{zafeiriou2011photoface}. }
	\begin{center}
		\begin{tabular}{c||c|c|c|c}
			\hline
			Method & Mean $\pm$ std  & $<20^{\circ}$ & $<25^{\circ}$ & $<30^{\circ}$   \\
			\hline\hline
			Pix2V~\cite{sela2017unrestricted}               & 33.9$\pm$5.6   & 24.8$\%$   & 36.1$\%$ & 47.6$\%$ \\
			Extreme~\cite{tran2018extreme}                  & 27.0$\pm$6.4   & 37.8$\%$   & 51.9$\%$ & 64.5$\%$ \\
			3DMM                                            & 26.3$\pm$10.2  & 4.3$\%$    & 56.1$\%$ & 89.4$\%$ \\
			3DDFA~\cite{zhu2016face}                        & 26.0$\pm$7.2   & 40.6$\%$   & 54.6$\%$ & 66.4$\%$ \\
			SfSNet~\cite{sengupta2018sfsnet}                & 25.5$\pm$9.3   & 43.6$\%$   & 57.5$\%$ & 68.7$\%$ \\
			PRN~\cite{feng2018joint}                        & 24.8$\pm$6.8   & 43.1$\%$   & 57.4$\%$ & 69.4$\%$ \\
			Cross-modal~\cite{abrevaya2020cross}            & 22.8$\pm$6.5   & 49.0$\%$   & 62.9$\%$ & 74.1$\%$ \\
			\hline \hline
			UberNet~\cite{kokkinos2017ubernet}          & 29.1$\pm$11.5  & 30.8$\%$   & 36.5$ \% $   & 55.2$\%$\\
			NiW~\cite{trigeorgis2017face}             & 22.0$\pm$6.3   & 36.6$\%$   & 59.8$ \% $   & 79.6$\%$\\
			Marr Rev~\cite{bansal2016marr}            & 28.3$\pm$10.1  & 31.8$\%$   & 36.5$ \% $   & 44.4$\%$\\
			SfSNet-ft~\cite{sengupta2018sfsnet}       & 12.8$\pm$5.4   & 83.7$\%$   & 90.8$\%$     & 94.5$\%$ \\
			LAP~\cite{zhang2021learning}       & 12.3$\pm$4.5   & 84.9$\%$   & 92.4$ \% $   & 96.3$\%$\\
			Cross-modal-ft~\cite{abrevaya2020cross}   & 12.0$\pm$5.3   & 85.2$\%$   & 92.0$ \% $   & 95.6$\%$\\
			HFN~\cite{wang2022towards}    & \textbf{11.3$\pm$7.7}  & \textbf{88.6}$\%$ & \textbf{94.4}$ \% $ & \textbf{97.2}$\%$\\	
			\hline \hline				
			Ours   & \textbf{10.1$\pm$6.5}  & \textbf{93.0}$\%$ & \textbf{96.8}$ \% $ & \textbf{98.4}$\%$\\						
			\hline
		\end{tabular}
	\end{center}
	\label{tb_phdb}
	\caption{Reconstruction error on the Florence dataset~\cite{bagdanov2011florence}. }
	\begin{center}
		\begin{tabular}{c||c|c|c|c}
			\hline
			Method & Mean $\pm$ std  & $<20^{\circ}$ & $<25^{\circ}$ & $<30^{\circ}$   \\
			\hline\hline
			Extreme~\cite{tran2018extreme}         & 19.2$\pm$2.2   & 64.7$\%$   & 51.9$\%$ & 64.5$\%$ \\
			SfSNet~\cite{sengupta2018sfsnet}       & 18.7$\pm$3.2   & 63.1$\%$   & 77.2$\%$ & 86.7$\%$ \\
			3DDFA~\cite{zhu2016face}               & 14.3$\pm$2.3   & 79.7$\%$   & 87.3$\%$ & 91.8$\%$ \\
			PRN~\cite{feng2018joint}               & 14.1$\pm$2.2   & 79.9$\%$   & 88.2$\%$ & 92.9$\%$ \\
			Cross-modal~\cite{abrevaya2020cross}   & 11.3$\pm$1.5   & 89.3$\%$   & 94.6$\%$ & 96.9$\%$ \\	
			HFN~\cite{wang2022towards}   & \textbf{10.1$\pm$3.4}  & \textbf{92.3}$\%$ & \textbf{95.6}$ \% $ & \textbf{97.8}$\%$\\	
			\hline\hline	
			Ours   & \textbf{9.8$\pm$3.2}  & \textbf{92.8}$\%$ & \textbf{96.1}$ \% $ & \textbf{98.3}$\%$\\																	
			\hline
		\end{tabular}
	\end{center}
	
	\label{tb_flor}
\end{table}	

\begin{figure*}[t]
	\begin{center}
		\begin{overpic}[scale=0.29]{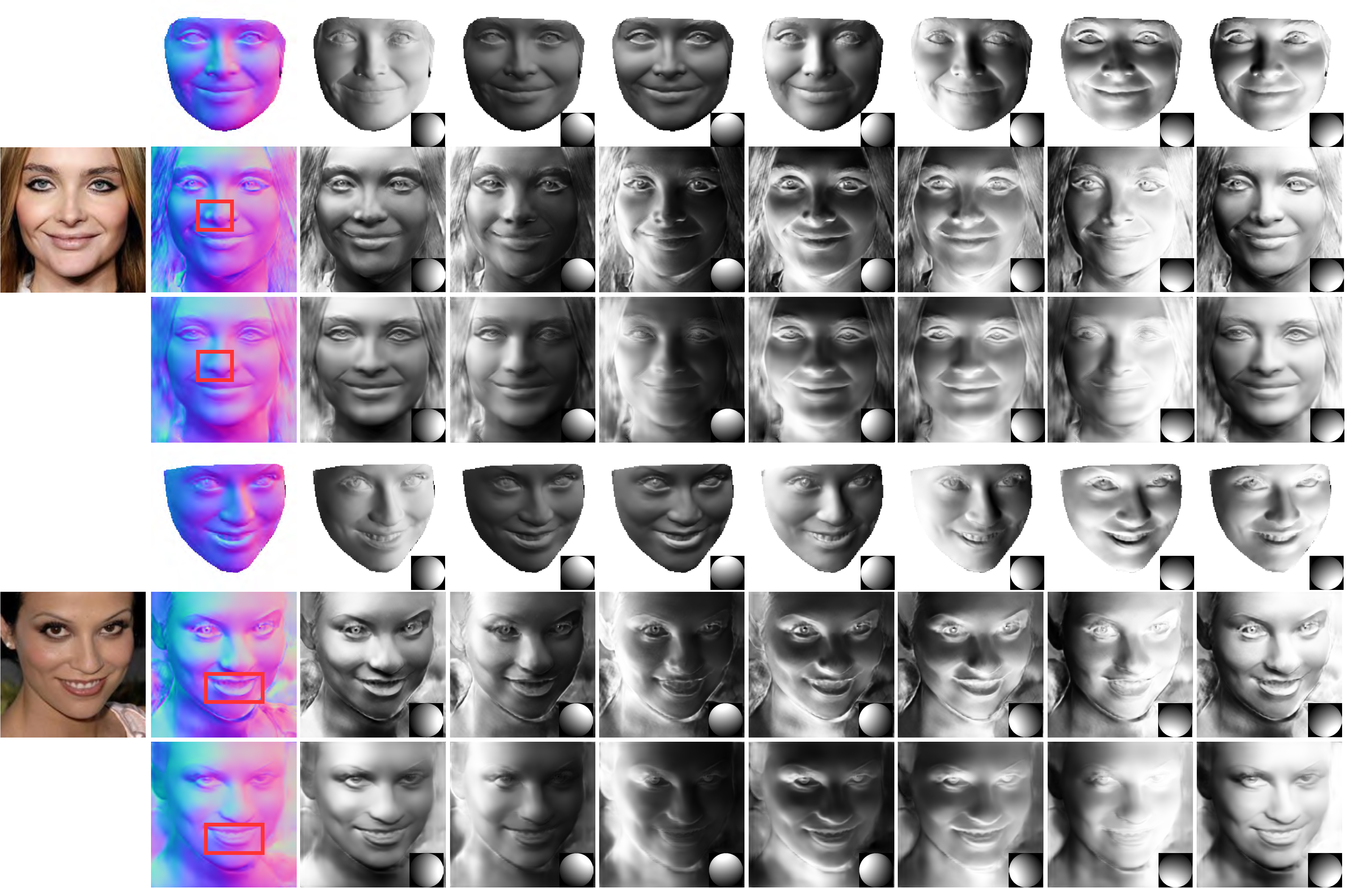}
			\put(2.5, -1.5){\footnotesize{Input}}
			\put(13, -1.5){\footnotesize{Normal}}
			\put(23, -1.5){\footnotesize{Shading 1}}
			\put(33.5, -1.5){\footnotesize{Shading 2}}
			\put(44.5, -1.5){\footnotesize{Shading 3}}
			\put(55.5, -1.5){\footnotesize{Shading 4}}
			\put(66, -1.5){\footnotesize{Shading 5}}
			\put(77, -1.5){\footnotesize{Shading 6}}
			\put(88, -1.5){\footnotesize{Shading 7}}
			
			\put(98, 60){ \footnotesize{\rotatebox{270}{CM}}}
			\put(98, 50){ \footnotesize{\rotatebox{270}{HFN}}}
			\put(98, 40){ \footnotesize{\rotatebox{270}{Ours}}}
			\put(98, 28){ \footnotesize{\rotatebox{270}{CM}}}
			\put(98, 18){ \footnotesize{\rotatebox{270}{HFN}}}
			\put(98, 7){ \footnotesize{\rotatebox{270}{Ours}}}
		\end{overpic}
		\caption{Comparison of normals and shading maps on the CelebA dataset~\cite{liu2015deep}. We generate shading maps using 7 different illumination directions to evaluate the normal accuracy of `CM'~\cite{abrevaya2020cross}, `HFN'~\cite{wang2022towards} and ours.}
		\label{com_hfn_cm_Vl7}
	\end{center}
\end{figure*}

Table~\ref{tb_phdb} presents a comparison between our methods (`Ours') and other state-of-the-art methods, including `Cross-modal-ft'~\cite{abrevaya2020cross}, `SfSNet-ft'\cite{sengupta2018sfsnet}, `Marr Rev'~\cite{bansal2016marr}, `NiW'\cite{trigeorgis2017face}, and `UberNet'~\cite{kokkinos2017ubernet}. All results were obtained using a cropped face image with a resolution of $256\times256$ pixels. The table reports the mean angular errors (in degrees) and the percentages of errors below $<20^{\circ}$, $<25^{\circ}$, and $<30^{\circ}$. Methods marked with `-ft' indicate that they were fine-tuned on the Photoface dataset~\cite{zafeiriou2011photoface}. Comparing our method to previous approaches trained on the Photoface, our method consistently achieves higher accuracy in normal estimation. The mean angular errors obtained by our method are significantly lower, indicating more accurate normal predictions. Additionally, the percentages of errors below different thresholds ($<20^{\circ}$, $<25^{\circ}$, $<30^{\circ}$) are consistently higher, indicating that our method produces normals with a higher level of precision.

\begin{figure}[t]
	\begin{center}
		\begin{overpic}[width=\linewidth]{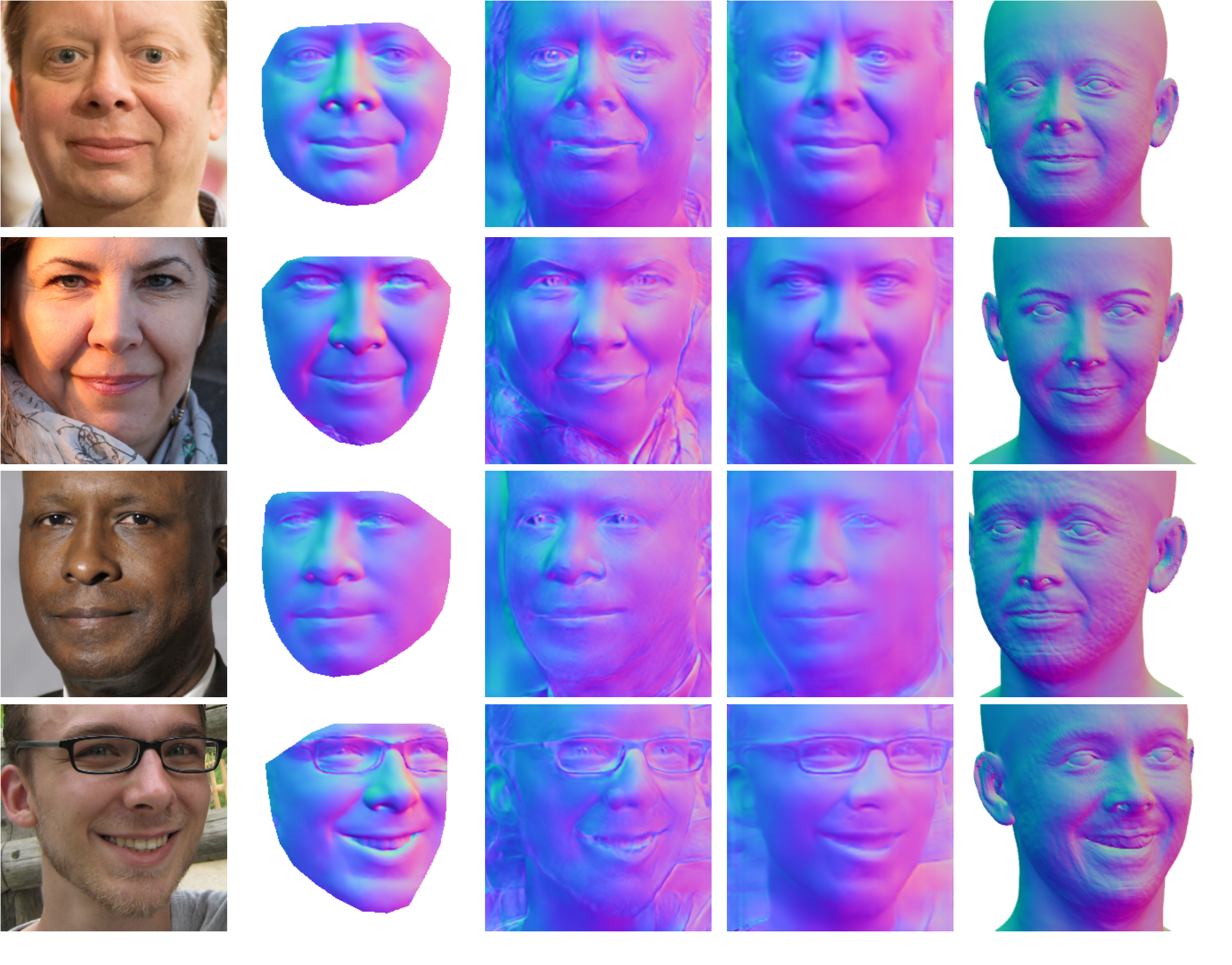}
			\put(6, -1){\footnotesize{Input}}
			\put(26, -1){\footnotesize{CM}}
			\put(46, -1){\footnotesize{HFN}}
			\put(66, -1){\footnotesize{Ours}}
			\put(83, -1){\footnotesize{EMOCA}}
		\end{overpic}
		\vspace{-6mm}
		\caption{Comparison of normal estimation results with the 3D face reconstruction method EMOCA \cite{danvevcek2022emoca}.}
		\label{comtog}
	\end{center}
\end{figure}
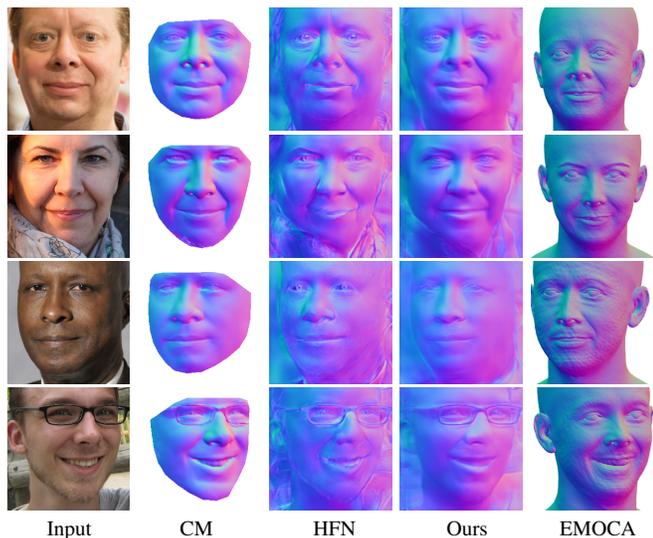
In Fig.\ref{F_errorMap}, we provide qualitative comparisons with the methods `HFN'~\cite{wang2022towards} and `CM'~\cite{abrevaya2020cross}. The figure showcases the normal estimations and normal error maps on test samples from the Photoface dataset~\cite{zafeiriou2011photoface}. The normal estimation errors are visualized using color maps, with darker colors indicating smaller errors (closer to 0 degrees). Our method produces more accuracy face normal estimations compared to `HFN' and `CM'. The normal estimations obtained by our method exhibit smaller errors, as evidenced by the darker and more localized regions in the error maps. This indicates that our method is able to capture more accurate and detailed normal information from the input face images, leading to improved normal estimation performance.

In Table~\ref{tb_flor}, we present the quantitative results on the Florence dataset~\cite{bagdanov2011florence}. To ensure a fair comparison, we follow the approach of previous work~\cite{abrevaya2020cross} and only compare methods using aligned output normals of face images. The table demonstrates that our proposed model outperforms the other methods, indicating its superior performance in handling out-of-distribution face images. This result further supports the effectiveness of our method in generalizing to unseen data and producing accurate normal estimations.
\begin{figure}[t]
	\begin{center}
		\begin{overpic}[width=\linewidth]{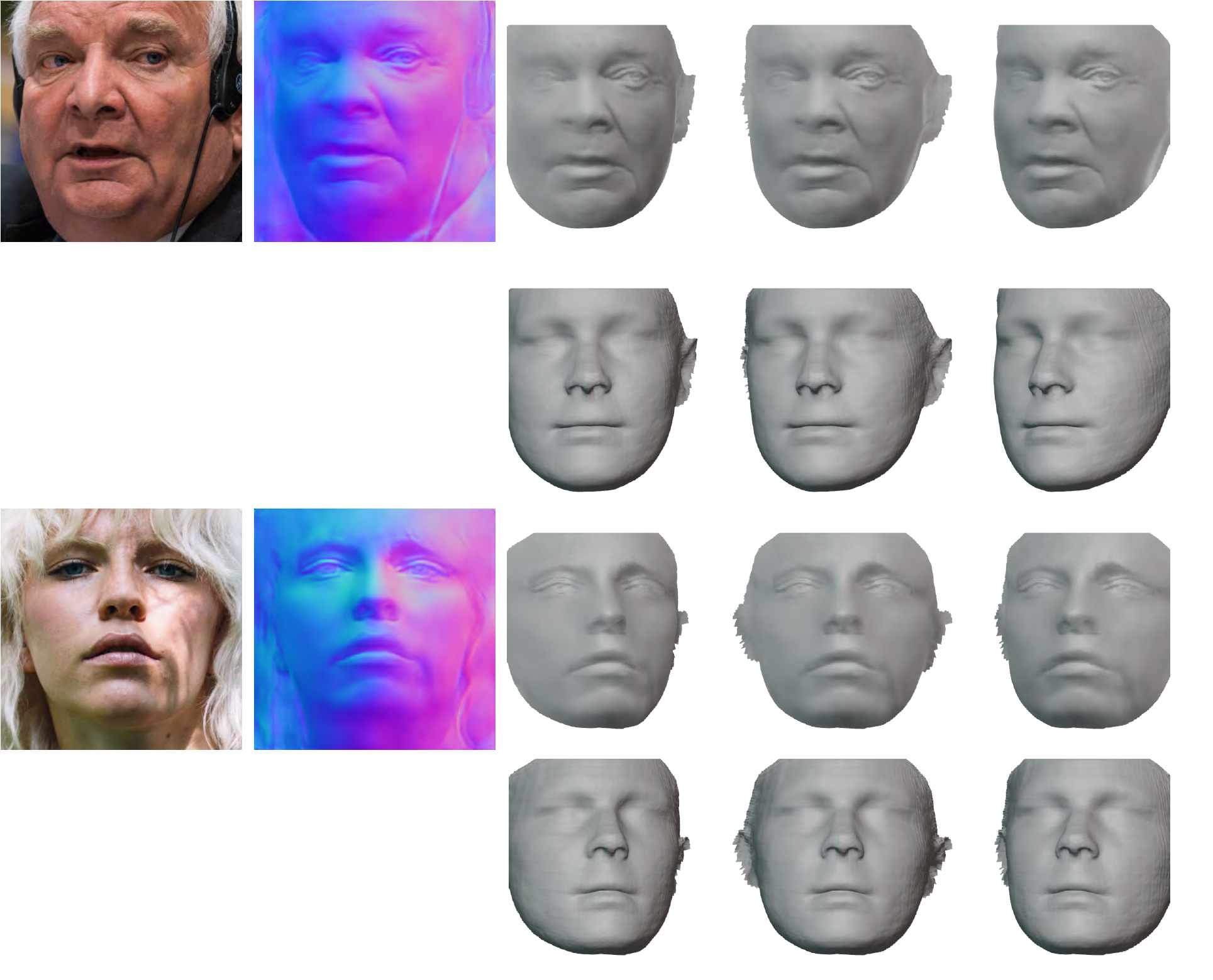}
			\put(97.2, 76){\footnotesize{\rotatebox{270}{Our-En}}}
			\put(97.2, 53){\footnotesize{\rotatebox{270}{PRN}}}
			\put(97.2, 35){\footnotesize{\rotatebox{270}{Our-En}}}
			\put(97.2, 15){\footnotesize{\rotatebox{270}{PRN}}}
		\end{overpic}
		\vspace{-25pt}
		\caption{Normal enhanced geometry details. PRN \cite{feng2018prn} is employed to estimate the coarse face geometry, `Ours-En' leverages normal mapping to further enhance the geometric details of the face. }
		\label{enhanceGeo}
	\end{center}
\end{figure}
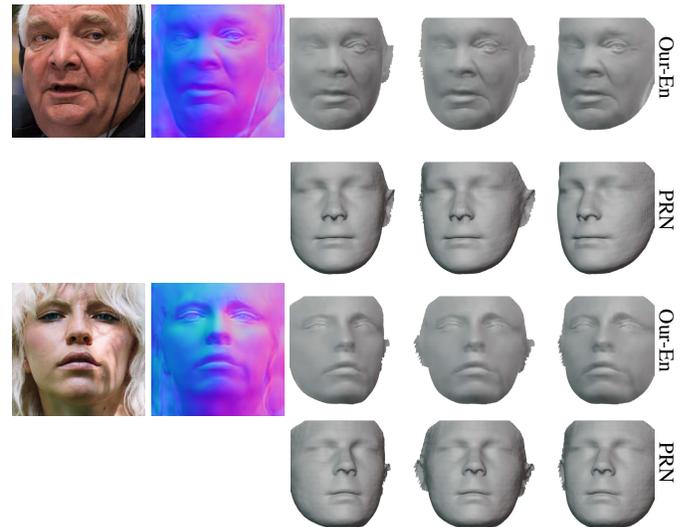

In Fig. \ref{com_hfn_cm_Vl7}, we show the shading maps generated by `CM' \cite{abrevaya2020cross}, `HFN' \cite{wang2022towards}, and `Ours' using seven different angular illuminations. The results indicate that the normal maps estimated by `CM' produce inaccurate shading effects (Shading 2 and Shading 3) when interacting with light. On the other hand, both `HFN' and `Ours' accurately depict the shading effects produced by various angles of light. Upon zooming in on shading maps, it is evident that `HFN' has some subtle artifacts (black spots on the nose or mouth), whereas `Ours' effectively removes these noises. This further confirms that Ours normals are more precise.

Fig. \ref{comtog} illustrates a comparison between our method and the normal produced by the state-of-the-art and representative face reconstruction algorithm, EMOCA \cite{danvevcek2022emoca}. As our approach is exclusively focused on the generation of facial normal maps, we compare the normal maps produced by the reconstructed facial model generated by EMOCA.
The results clearly demonstrate that our method excels at recovering intricate and detailed facial normal, while the normal map generated by the EMOCA method lacks crucial facial details.

\begin{figure}[t]
	\begin{center}
		\begin{overpic}[width=\linewidth]{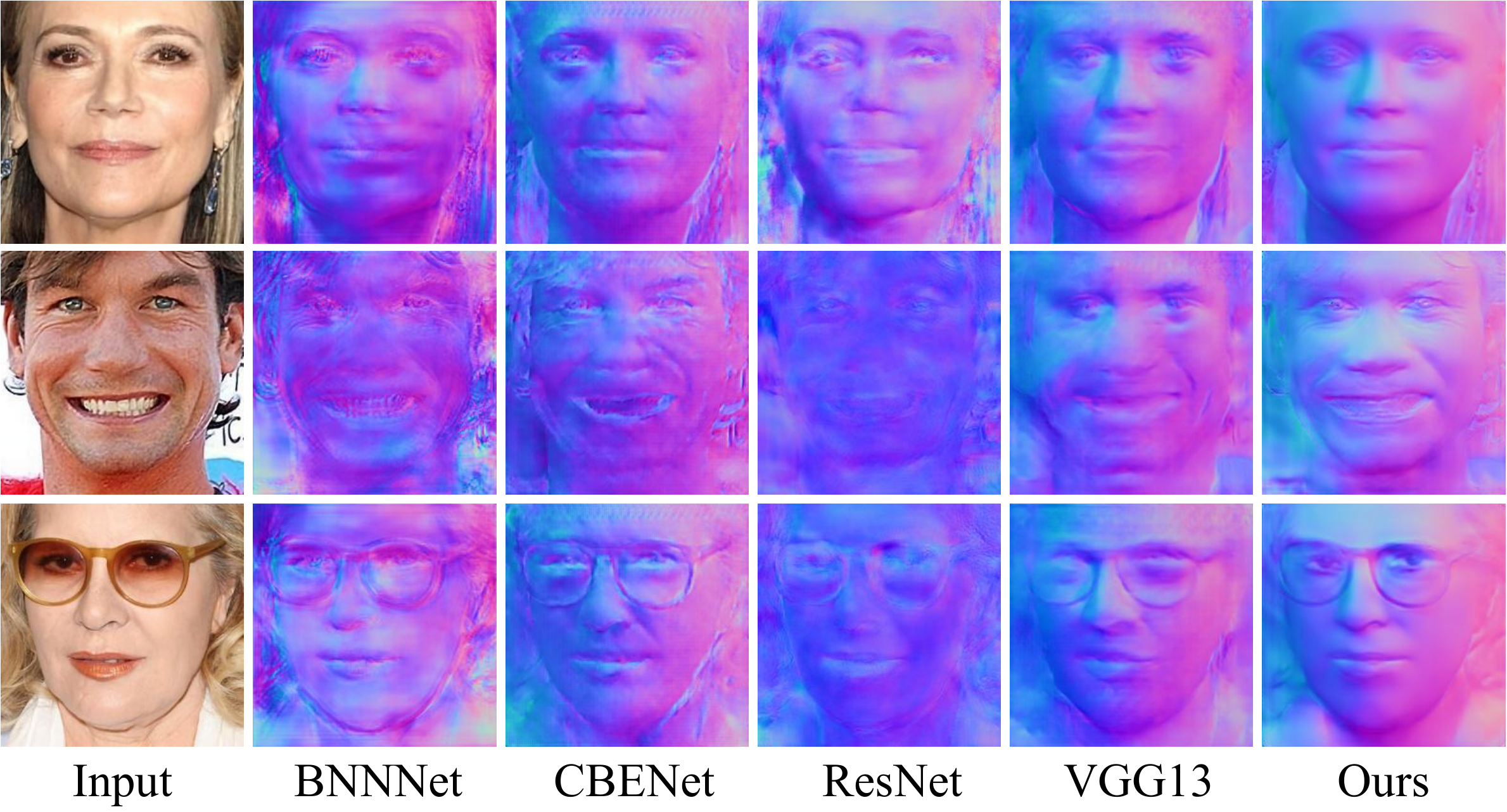}
		\end{overpic}
		\vspace{-8mm}
		\caption{Comparison results for out-of-distribution using different networks on the CelebA dataset~\cite{liu2015deep}. The last layer of BNNNet~\cite{li2023spatially}, CBENet~\cite{zhang2023document}, ResNet~\cite{zhu2017unpaired} and VGG13~\cite{simonyan2014very} has been modified with a tanh function, as described in the cited papers.}
		\label{ablationNet}
	\end{center}
\end{figure}
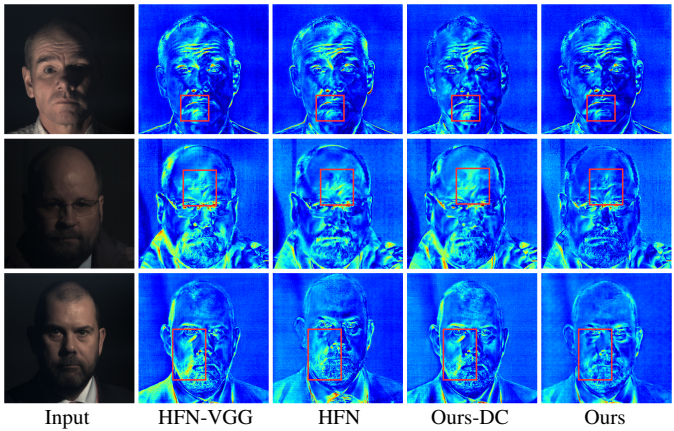
\begin{figure}[t]
	\begin{center}
		\begin{overpic}[width=\linewidth]{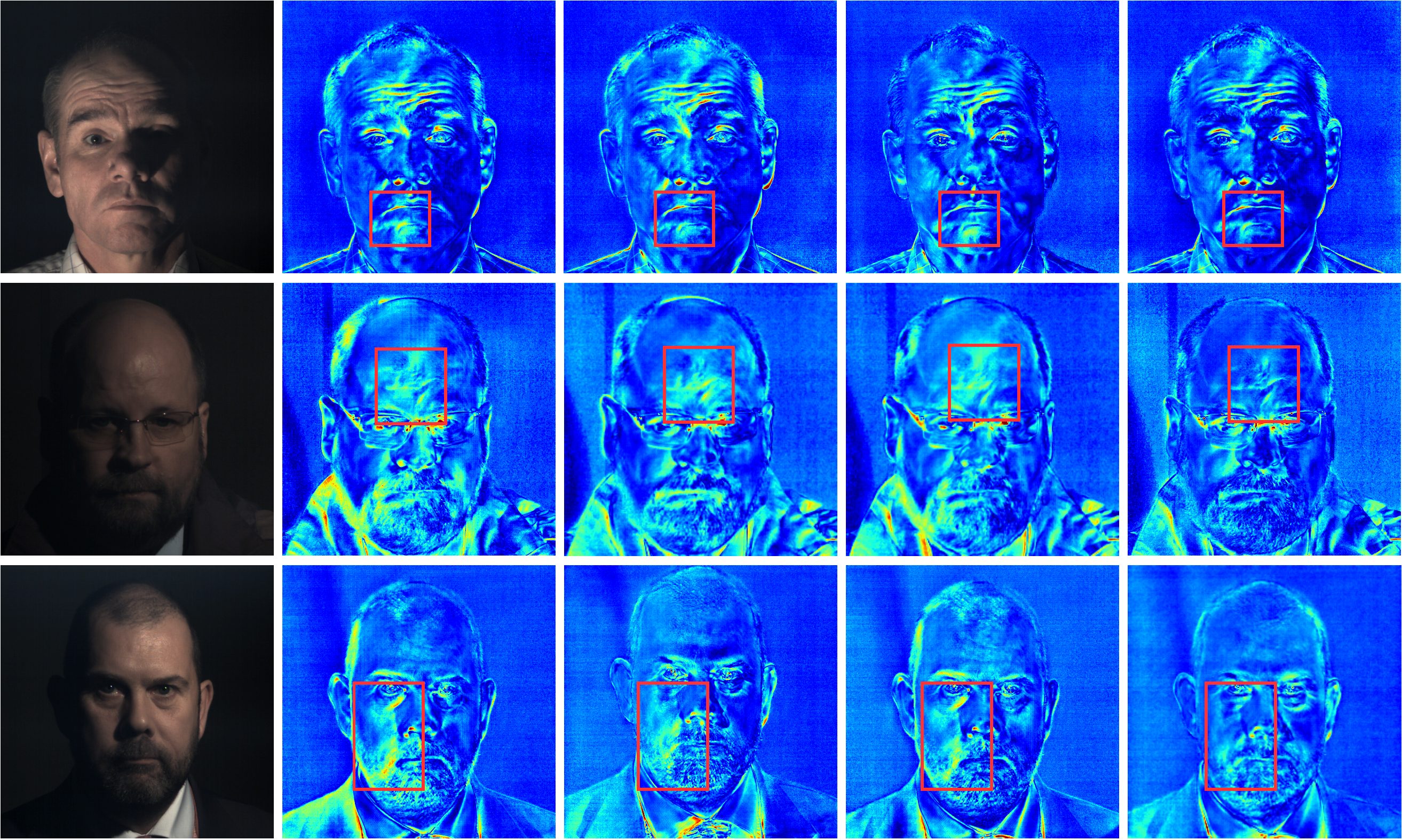}
			\put(6, -3){\footnotesize{Input}}
			\put(23, -3){\footnotesize{HFN-VGG}}
			\put(47, -3){\footnotesize{HFN}}
			\put(64, -3){\footnotesize{Ours-DC}}
			\put(87, -3){\footnotesize{Ours}}
		\end{overpic}
		\caption{The comparison on different normal feature encoder networks ($E_{R}$). `HFN-VGG' represents the model trained with `HFN' \cite{wang2022towards} based $E_{R}$ without using VGG features, `Ours-DC' uses a double convolution layer followed by a pooling layer.}
		\label{NormalEncoderCom}
	\end{center}
\end{figure}
Normal mapping is a technique widely used in computer graphics to enhance the visual detail of a 3D model without increasing its polygon count. By mapping a detailed normal map onto the surface of a model, the appearance of intricate surface details such as wrinkles and beards can be convincingly simulated. Therefore, we map the estimated normal map to the face geometry model for enhancing the face details, and the result is shown in Fig \ref{enhanceGeo}. From the figure, it can be seen that the facial geometry details can be enhanced by means of normal mapping.

\section{Ablation studies}
\label{AbS}

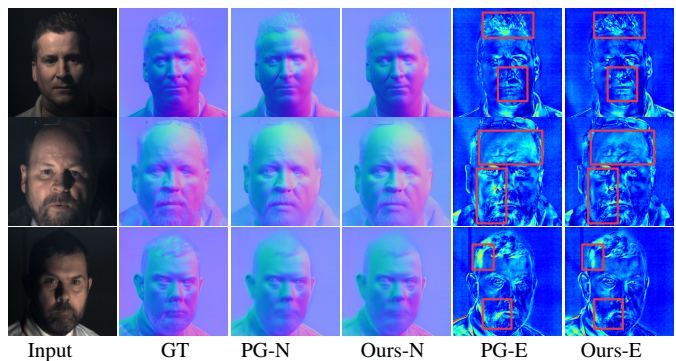
\begin{figure}[t]
	\begin{center}
		\begin{overpic}[width=\linewidth]{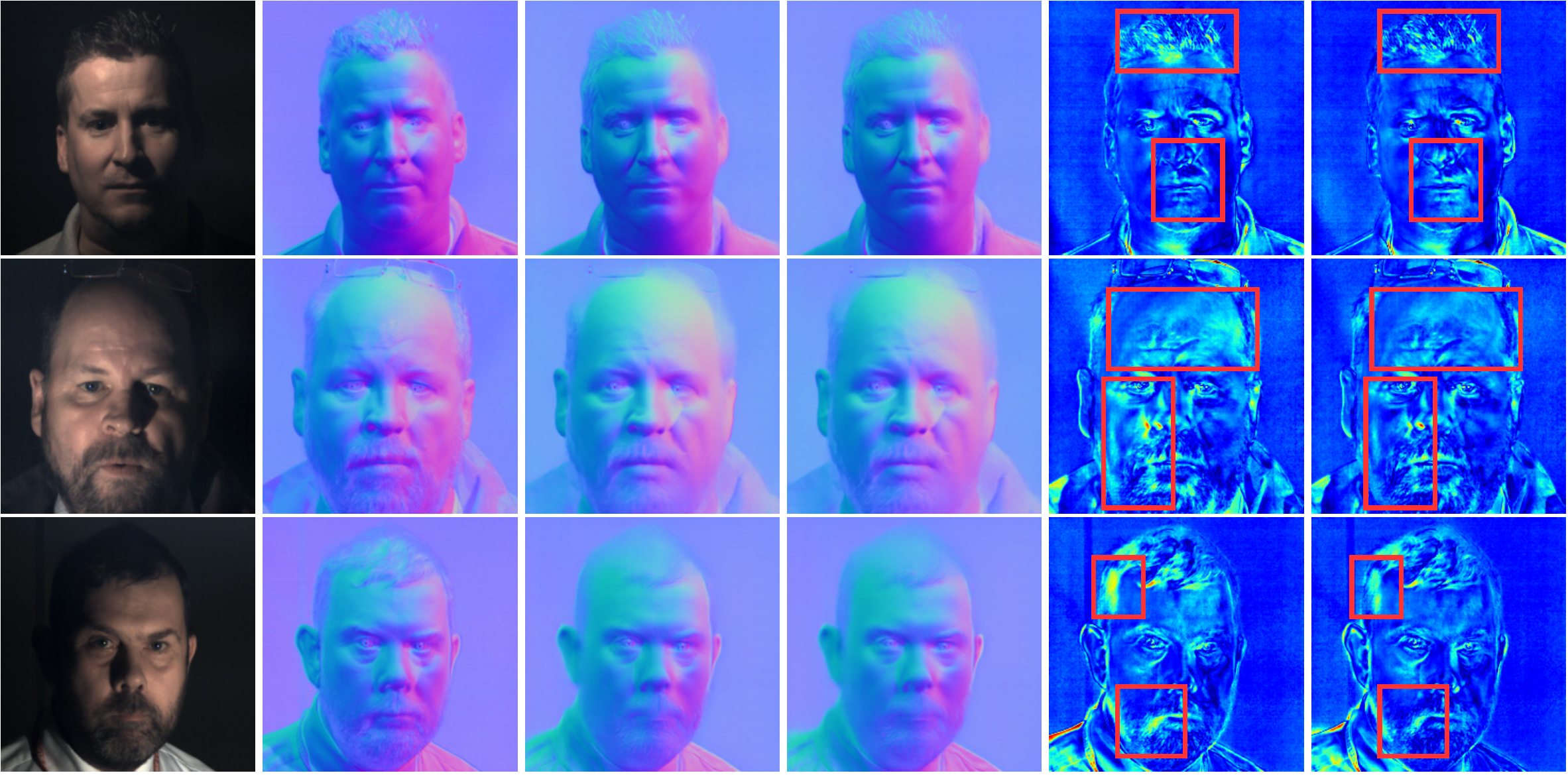}
			\put(3, -3){\footnotesize{Input}}
			\put(23, -3){\footnotesize{GT}}
			\put(35, -3){\footnotesize{PG-N}}
			\put(53, -3){\footnotesize{Ours-N}}
			\put(71, -3){\footnotesize{PG-E}}
			\put(86, -3){\footnotesize{Ours-E}}
		\end{overpic}
		\caption{Normal comparison with different discriminators on the Photoface~\cite{zafeiriou2011photoface}. `GT', `PG', `-N', and `-E' are ground truths, results produced by Patch-based discriminator \cite{isola2017image} model, coarse normals, refined normals and normal error maps, respectively.}
		\label{errorMap512}
	\end{center}
\end{figure}
\textbf{Generalization ability.} Fig. \ref{ablationNet} and Table \ref{NormalEncoder_COM} demonstrate our testing of the performance of different networks by training end-to-end normal estimation methods using commonly used classical network structures. Although models trained with paired data under limited conditions can achieve convergence on the training data and obtain good evaluation metrics, their model performance capability is poor when tested on in-the-wild face images. However, we are able to perform secondary refinement of the normals using limited data conditions to give the model strong generalization capabilities.

\begin{table}[t]
	\caption{Comparison in normal reconstruction error with different configurations on the Photoface dataset~\cite{zafeiriou2011photoface}.}
	\begin{center}
		\begin{tabular}{c||c|c|c|c}
			\hline
			Experiments & Mean $\pm$ std  & $<20^{\circ}$ & $<25^{\circ}$ & $<30^{\circ}$   \\
			\hline\hline
			
			BNNNet~\cite{li2023spatially}       & 11.7$\pm$8.8    & 86.2$\%$   & 92.4$\%$ & 95.5$\%$ \\	
			
			CBENet~\cite{zhang2023document}       & 10.5$\pm$7.4    & 90.6$\%$   & 95.3$\%$ & 97.5$\%$ \\	
			
			ResNet~\cite{zhu2017unpaired}       & 7.9$\pm$5.4    & 96.1$\%$   & 97.9$\%$ & 98.8$\%$ \\	
			
			VGG13~\cite{simonyan2014very}       & 10.1$\pm$8.2    & 89.8$\%$   & 94.1$\%$ & 96.5$\%$ \\	
			
			HFN-VGG~\cite{wang2022towards}        & 11.4$\pm$7.9   & 88.1$\%$   & 94.1$\%$ & 96.8$\%$ \\
			
			HFN~\cite{wang2022towards}       & 11.3$\pm$7.7    & 88.6$\%$   & 94.4$\%$ & 97.2$\%$ \\	
			Ours-DC   & 11.1$\pm$7.2  & 91.6$\%$ & 95.3$ \% $ & 98.1$\%$\\			
			Ours   & \textbf{10.1$\pm$6.5}  & \textbf{93.0}$\%$ & \textbf{96.8}$ \% $ & \textbf{98.4}$\%$\\																
			\hline
		\end{tabular}
	\end{center}
	\label{NormalEncoder_COM}
\caption{Comparative experiments with existing methods when reliable references are unavailable. `NoTP' stands for a number of training pairs used during our training. }
	\begin{center}
		\begin{tabular}{c||c|c|c|c|c}
			\hline
			Experiments    & NoTP     & Mean $\pm$ std  & $<20^{\circ}$ & $<25^{\circ}$ & $<30^{\circ}$   \\
			\hline\hline
   		SfSNet         &250k     & 12.8$\pm$5.4   & 83.7$\%$   & 90.8$\%$    & 94.5$\%$ \\
			Cross-modal    &21k      & 11.2$\pm$7.2   & 92.1$\%$   & 95.2$\%$    & 97.1$\%$ \\
			Ours (25\%)    &2.3k     & 11.1$\pm$6.8   & 92.3$\%$   & 95.7$\%$    & 97.4$\%$ \\
			\hline\hline   
			Ours (75\%)    &7.1k     & 10.4$\pm$7.1   & 92.7$\%$   & 96.1$\%$    & 98.1$\%$ \\
			Ours (100\%)   &9.5k     & 10.1$\pm$6.5   & 93.0$\%$   & 96.8$\%$    & 98.4$\%$ \\										
			\hline
		\end{tabular}
	\end{center}
	\label{Table:trainNumbers}
\end{table}	

\textbf{Normal feature encoder.}  In order to evaluate the effectiveness of VGG feature extraction in $E_{R}$, we conducted a comparison between `HFN-VGG' \cite{wang2022towards} without using pretrained VGG, `HFN', `Ours-DC' only with a double convolution followed by a pooling layer and `Ours'. The results of this comparison are presented in Fig. \ref{NormalEncoderCom} and Table \ref{NormalEncoder_COM}, which show the estimated normals are minimally influenced by the network structure. Specifically, even without the VGG module, the `HFN' model still produced decent estimated normal maps. This suggests that a smaller network can effectively extract normal distribution features of size 256 for the purpose of normal refinement, as demonstrated by `Ours-DC'. These findings imply that the VGG feature extraction module can be replaced with a simpler and more compact architecture. In this paper, we utilize the FPN-based structure to extract normal features while ensuring that we do not lose too much information. By utilizing the FPN network structure, it is possible to not only significantly decrease the number of model parameters but also guarantee the extraction of accurate features at varying scales. 

\textbf{Self-attention module.} We further investigated the effectiveness of the attention mechanism by training an additional model using PatchGAN's discriminator \cite{isola2017image}. The comparison between the attention mechanism and the PatchGAN-based discriminator is shown in Fig. \ref{errorMap512}. The results clearly indicate that the attention mechanism enables our model to capture detailed information from the face image, highlighting its efficacy. This finding also suggests that the coarse samples generated in the first stage, utilizing the attention mechanism, can effectively serve as a reliable input for the normal refinement in the second phase of our framework.

\begin{figure}[t]
	\begin{center}
		\begin{overpic}[width=\linewidth]{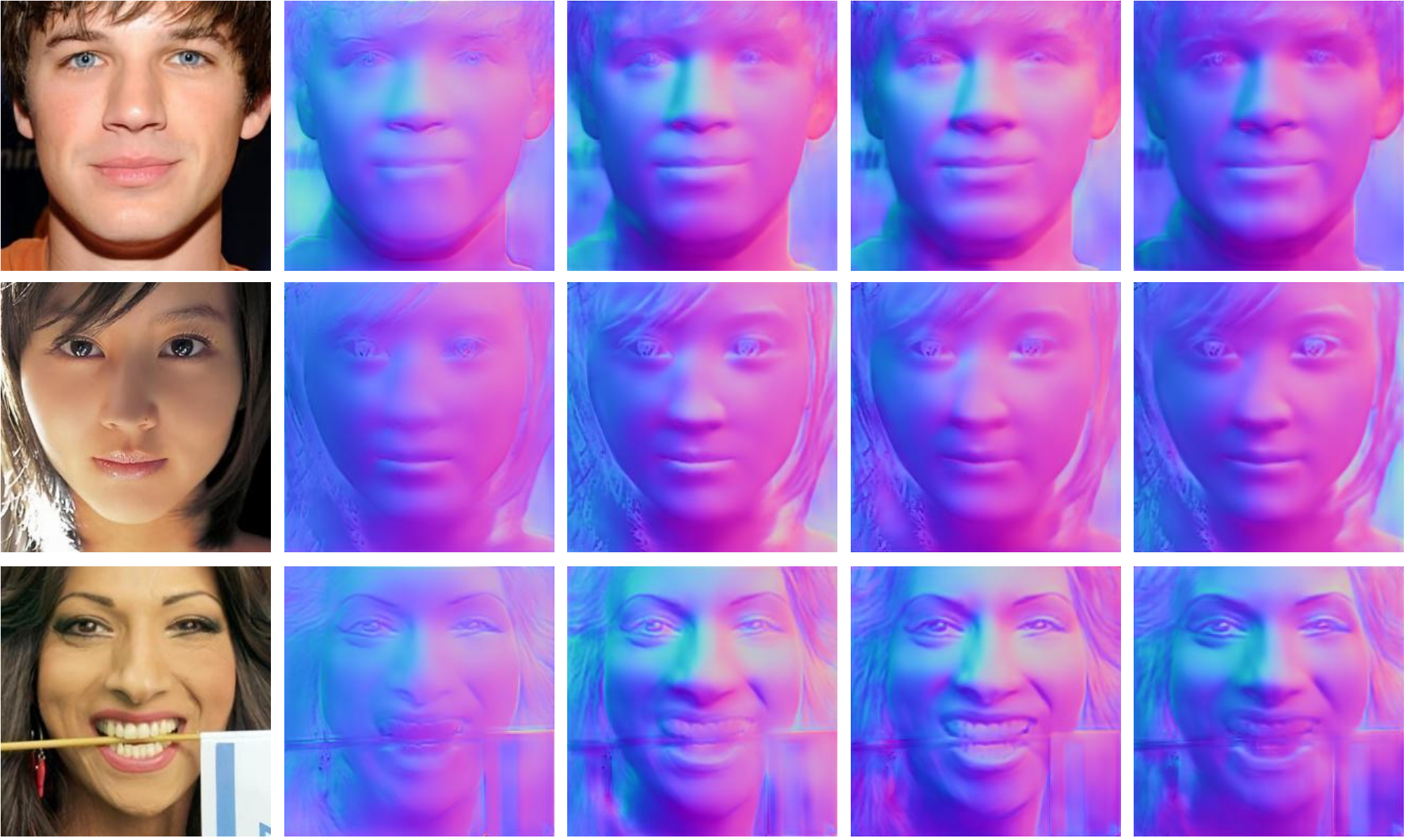}
			\put(4.5, -3){\footnotesize{Input}}
			\put(28, -3){\footnotesize{25$\%$}}
			\put(48, -3){\footnotesize{50$\%$}}
			\put(68, -3){\footnotesize{75$\%$}}
			\put(87, -3){\footnotesize{Ours}}
		\end{overpic}
		\caption{Normal results with different initializations on the FFHQ dataset~\cite{karras2019style}. Here, we can find that the coarse prediction model learns the distribution of normal roughly, and our method can obtain high-fidelity normal with different iterations in the refined stage.}
		\label{dataCountCeleba}
	\end{center}
	\begin{center}
		\begin{overpic}[width=\linewidth]{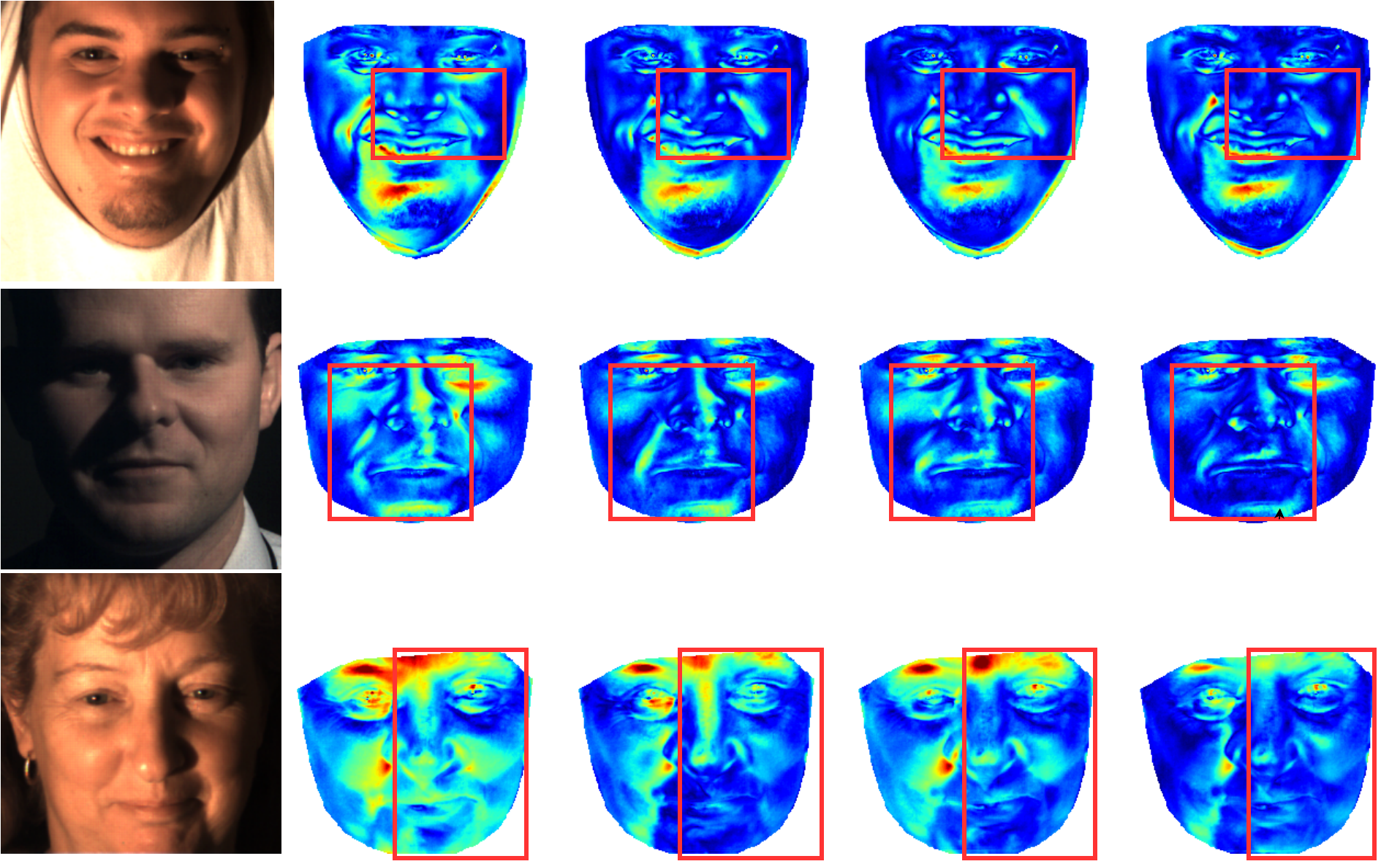}
			\put(5, -3){\footnotesize{Input}}
			\put(28, -3){\footnotesize{25$\%$}}
			\put(48, -3){\footnotesize{50$\%$}}
			\put(68, -3){\footnotesize{75$\%$}}
			\put(87, -3){\footnotesize{Ours}}
		\end{overpic}
		\caption{Normal error maps with different initializations on the Photoface dataset~\cite{zafeiriou2011photoface}.}
		\label{dataCountPhoto}
	\end{center}
\end{figure}
\textbf{Robustness to varying amounts of training data.} To evaluate the robustness of our method to variations in training data, we conducted experiments by dividing the training set and randomly selecting 25\%, 50\%, and 75\% of the data for model training. The results, shown in Fig. \ref{dataCountCeleba} and Fig. \ref{dataCountPhoto}, indicate that even with a reduced data volume of 25\%, our method performs well not only on data consistent with the training data but also on data outside the training data distribution. To provide a more intuitive comparison of the effects of different data amounts, we present quantitative evaluation results in Table \ref{Table:trainNumbers}. From the table, it can be observed that our proposed method exhibits a certain level of robustness in accurately estimating face normals, even under conditions of limited training data.

\begin{figure}[t]
	\begin{center}
		\begin{overpic}[width=\linewidth]{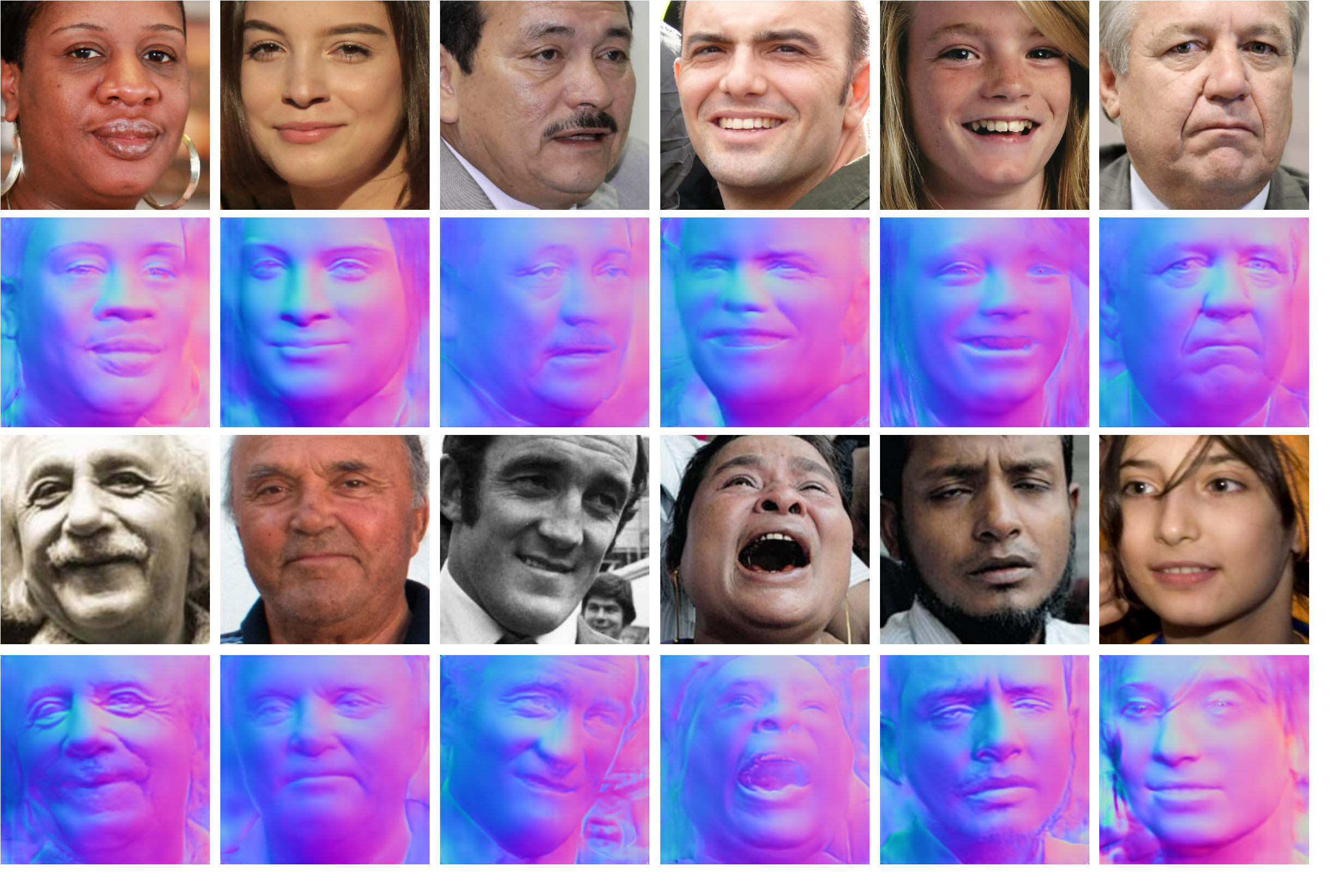}
			\put(98.2, 63){\rotatebox{270}{FFHQ}}
			\put(98.2, 48){\rotatebox{270}{Normal}}
			\put(98.2, 32){\rotatebox{270}{300-W}}
			\put(98.2, 15.3){\rotatebox{270}{Normal}}
		\end{overpic}
		\vspace{-25pt}
		\caption{Normal results on the FFHQ~\cite{karras2019style} and 300-W \cite{sagonas2013300}. Despite not having seen such faces during training, our model generalizes reasonably to these images.}
		\label{ffhqmore}
	\end{center}
	\begin{center}
		\begin{overpic}[width=\linewidth]{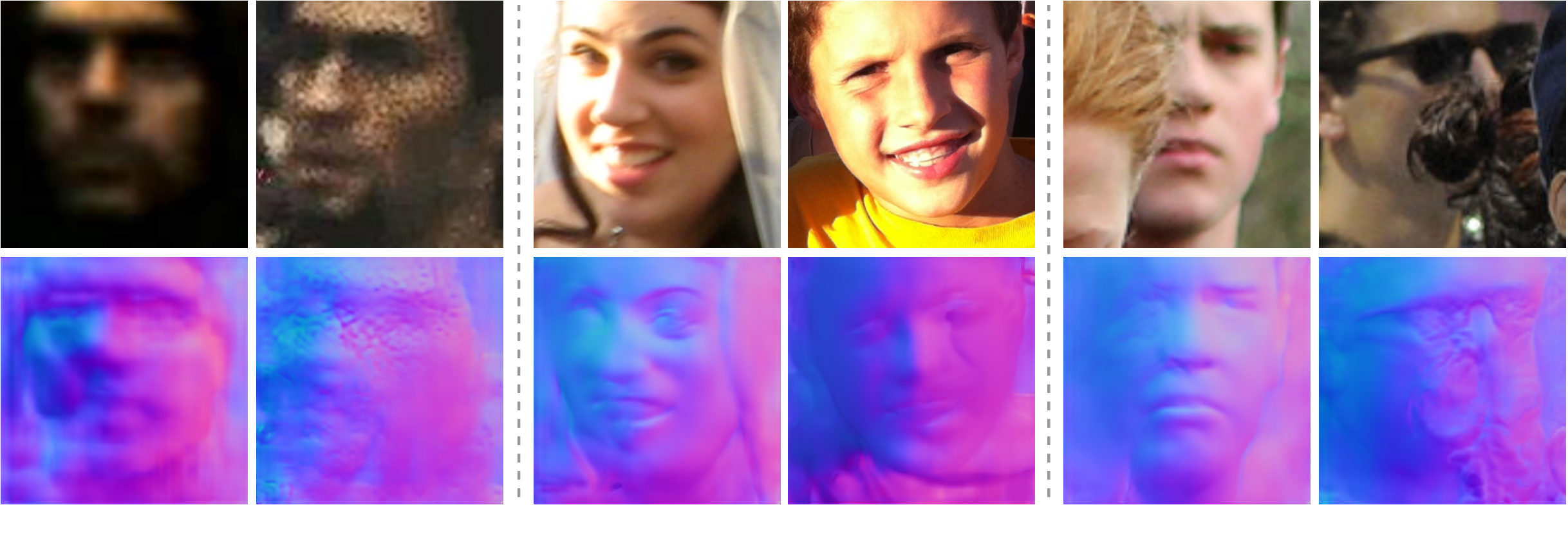}
			\put(14, -1){\footnotesize{(a)}}
			\put(50, -1){\footnotesize{(b)}}
			\put(86, -1){\footnotesize{(c)}}
		\end{overpic}
		\vspace{-17pt}
		\caption{Results for low-quality images (a), extreme lighting conditions (b), and faces that are partially covered (c).}
		\label{Limitation}
	\end{center}
\end{figure}

To validate the model, we tested our model on the challenging dataset, FFHQ~\cite{karras2019style} and 300-W \cite{sagonas2013300}. As shown in Fig. \ref{ffhqmore}, our model performs well on those datasets and is particularly effective in recovering finer-grained face normals. This is evident in regions where the face geometry changes, such as near the corners of wrinkled eyes. These results highlight the robustness and accuracy of our model in capturing detailed facial geometry.

\textbf{Running time.}  We compare our model with `CM' \cite{abrevaya2020cross} and `HFN' \cite{wang2022towards} in running time to verify the efficiency of our work in Table \ref{para_flops}. The table shows that the `CM' has clear advantages in terms of model parameters and FLOPs. However, this method requires intricate and complex network structures. Furthermore, it performs worse than `HFN' and `Ours' in terms of normal estimation accuracy. In comparison to `HFN', `Ours' is capable of not only reducing the model parameters to some extent but also significantly simplifying the model while improving normal estimation accuracy.
\begin{table}[t]
	\caption{Comparison of running times using a $256 \times 256$ resolution image, all evaluated on the same machine.}
	\begin{center}
		\begin{tabular}{c||c|c|c}
			\hline
			Method & Parameters(M)  & FLOPs(G) & Running time(ms)  \\
			\hline\hline
			
			CM~\cite{abrevaya2020cross}   & 35.2    &  49    &  10  \\	
			
			HFN~\cite{wang2022towards}    & 126.4   &  152   &  17  \\	
			
			Ours                          & 82.9    &  90    &  13  \\	
			
			\hline
		\end{tabular}
	\end{center}
	\label{para_flops}
\end{table}





%

\section{Concluding Remarks}
In this paper, we proposed a novel framework for high-fidelity face normal estimation. Our method draws inspiration from exemplar-based learning and leverages a coarse exemplar normal as guidance to generate final fine-grained results. The key strength of our approach lies in the conversion of the coarse exemplar normal into normal features and refinement through feature modulation. This mechanism can not only boost the estimation both qualitatively and quantitatively, but also grant it strong generalization capabilities, allowing our method to effectively handle out-of-distribution face images. Comprehensive qualitative and quantitative evaluations have been conducted to demonstrate that our method outperforms state-of-the-art approaches by a large margin.

While our method shows its robustness in many challenging scenarios, such as faces with wrinkles and beards, there are still failure cases, as illustrated in Fig.\ref{Limitation}. Images of very low quality (Fig.\ref{Limitation} (a)) and those with extreme lighting conditions or shading (Fig.\ref{Limitation} (c)) can lead to inaccurate normal reconstructions. Additionally, our method struggles with occluded face images, as shown in Fig.\ref{Limitation} (d)). We recognize that these unrestricted scenarios pose challenges to our method and also most existing approaches. It is desirable to explore more advanced variations to address these challenging and unrestricted cases more effectively.


\ifCLASSOPTIONcaptionsoff
\newpage
\fi



%

\bibliographystyle{ieee}
\bibliography{ref}

\end{document}